\documentclass[acmtog,screen,nonacm]{acmart}
\usepackage{booktabs}
\usepackage{graphicx}
\usepackage{wrapfig}
\usepackage{algorithm}
\usepackage[noend]{algpseudocode}
\usepackage{amsmath}
\usepackage{enumitem}

\renewcommand{\paragraph}[1]{\noindent\textbf{#1.}}

\newcommand{\rev}[1]{\textcolor{black}{#1}}

\newcommand{\figref}[1]{Fig.~\ref{#1}}

\usepackage{xspace}
\makeatletter
\DeclareRobustCommand\onedot{\futurelet\@let@token\@onedot}
\def\@onedot{\ifx\@let@token.\else.\null\fi\xspace}

\makeatother

\newcommand{\x}{\mathbf{x}}

\newcommand{\bC}{\mathbf{C}}

\newcommand{\bE}{\mathbf{E}}
\newcommand{\bF}{\mathbf{F}}

\newcommand{\bK}{\mathbf{K}}

\newcommand{\bS}{\mathbf{S}}

\newcommand{\bV}{\mathbf{V}}

\newcommand{\bff}{\mathbf{f}}

\newcommand{\bp}{\mathbf{p}}

\newcommand{\bs}{\mathbf{s}}

\newcommand{\bu}{\mathbf{u}}

\newcommand{\bx}{\mathbf{x}}

\newcommand{\ssm}{\mathcal{M}}
\newcommand{\holo}{\mathcal{P}}

\usepackage{amsmath}

\DeclareMathOperator*{\argmin}{arg\,min}

\newcommand{\srefhomo}{Appendix~\ref{sec:homogen}}
\newcommand{\sreftraining}{Appendix~\ref{sec:imple_detail}}
\newcommand{\srefHeterdesign}{Appendix~\ref{sec:heterogeneous}}
\newcommand{\srefmorevis}{Appendix~\ref{sec:more_vis}}

\providecommand{\E}{\mathbb{E}}

\copyrightyear{2025}
\acmYear{2025}
\setcopyright{cc}
\setcctype{by}
\acmDOI{10.1145/3721238.3730682}

\authorsaddresses{}
\newcommand{\mindauthor}[2]{\mbox{#1\textsuperscript{#2}}}
\makeatletter
\renewcommand{\@mkauthors}{%
  \gdef\@currentauthors{}%
  \global\setbox\mktitle@bx=\vbox{%
    \unvbox\mktitle@bx
    \centering\normalfont\large
    \mindauthor{Tianyang Xue}{1}\quad
    \mindauthor{Longdu Liu}{1}\quad
    \mindauthor{Lin Lu}{1,*}\quad
    \mindauthor{Paul Henderson}{2}\quad
    \mindauthor{Pengbin Tang}{3}\par
    \mindauthor{Haochen Li}{1}\quad
    \mindauthor{Jikai Liu}{1}\quad
    \mindauthor{Haisen Zhao}{1}\quad
    \mindauthor{Hao Peng}{4}\quad
    \mindauthor{Bernd Bickel}{3}\par\smallskip
    \small
    \textsuperscript{1}Shandong University, China\quad
    \textsuperscript{2}University of Glasgow, United Kingdom\par
    \textsuperscript{3}ETH Zurich, Switzerland\quad
    \textsuperscript{4}CrownCAD, China\par\medskip
  }%
}
\renewcommand{\@copyrightpermission}{%
  Author version with supplementary materials.
  Published in \emph{SIGGRAPH Conference Papers '25}.
  \href{https://doi.org/10.1145/3721238.3730682}{doi:10.1145/3721238.3730682}.\par
  \textcopyright~2025 Copyright held by the owner/author(s).
  Licensed under \href{https://creativecommons.org/licenses/by/4.0/}{CC BY 4.0}.%
}
\makeatother

\begin{document}
\title{MIND: Microstructure INverse Design with Generative Hybrid Neural Representation}

\author{Tianyang Xue}
\email{TimHsue@gmail.com}
\affiliation{
  \institution{Shandong University}
  \city{Qingdao}
  \country{China}
}

\author{Longdu Liu}
\email{liulongdu@163.com}
\affiliation{
  \institution{Shandong University}
  \city{Qingdao}
  \country{China}
}

\author{Lin Lu}
\email{llu@sdu.edu.cn}
\affiliation{
  \institution{Shandong University}
  \city{Qingdao}
  \country{China}
}
\authornote{Corresponding author: Lin Lu (\href{mailto:llu@sdu.edu.cn}{llu@sdu.edu.cn}).}

\author{Paul Henderson}
\email{paul@pmh47.net}
\affiliation{
  \institution{University of Glasgow}
  \city{Glasgow}
  \country{United Kingdom}
}

\author{Pengbin Tang}
\email{tangpengbin@gmail.com}
\affiliation{
  \institution{ETH Zurich}
  \city{Zurich}
  \country{Switzerland}
}

\author{Haochen Li}
\email{lihaochen@mail.sdu.edu.cn}
\affiliation{
  \institution{Shandong University}
  \city{Qingdao}
  \country{China}
}

\author{Jikai Liu}
\email{jikai_liu@sdu.edu.cn}
\affiliation{
  \institution{Shandong University}
  \city{Jinan}
  \country{China}
}

\author{Haisen Zhao}
\email{haisenzhao@sdu.edu.cn}
\affiliation{
  \institution{Shandong University}
  \city{Qingdao}
  \country{China}
}

\author{Hao Peng}
\email{penghao@crowncad.com}
\affiliation{
  \institution{CrownCAD}
  \city{Jinan}
  \country{China}
}

\author{Bernd Bickel}
\email{bickelb@ethz.ch}
\affiliation{
  \institution{ETH Zurich}
  \city{Zurich}
  \country{Switzerland}
}

\renewcommand{\shortauthors}{Xue et al.}

\begin{abstract}

The inverse design of microstructures plays a pivotal role in optimizing metamaterials with specific, targeted physical properties. While traditional forward design methods are constrained by their inability to explore the vast combinatorial design space, inverse design offers a compelling alternative by directly generating structures that fulfill predefined performance criteria. However, achieving precise control over both geometry and material properties remains a significant challenge due to their intricate interdependence. Existing approaches, which typically rely on voxel or parametric representations, often limit design flexibility and structural diversity.

In this work, we present a novel generative model that integrates latent diffusion with Holoplane, an advanced hybrid neural representation that simultaneously encodes both geometric and physical properties. This combination ensures superior alignment between geometry and properties. Our approach generalizes across multiple microstructure classes, enabling the generation of diverse, tileable microstructures with significantly improved property accuracy and enhanced control over geometric validity, surpassing the performance of existing methods.
We introduce a multi-class dataset encompassing a variety of geometric morphologies, including truss, shell, tube, and plate structures, to train and validate our model. Experimental results demonstrate the model’s ability to generate microstructures that meet target properties, maintain geometric validity, and integrate seamlessly into complex assemblies. Additionally, we explore the potential of our framework through the generation of new microstructures, cross-class interpolation, and the infilling of heterogeneous microstructures.
Code and data for this paper are at \href{https://github.com/TimHsue/MIND}{https://github.com/TimHsue/MIND}.

\end{abstract}

\ccsdesc[500]{Computing methodologies~Shape modeling}
\ccsdesc[300]{Computing methodologies~Graphics systems and interfaces}

\keywords{microstructures, generative design, neural networks, additive manufacturing}

\begin{teaserfigure}
  \includegraphics[width=\textwidth]{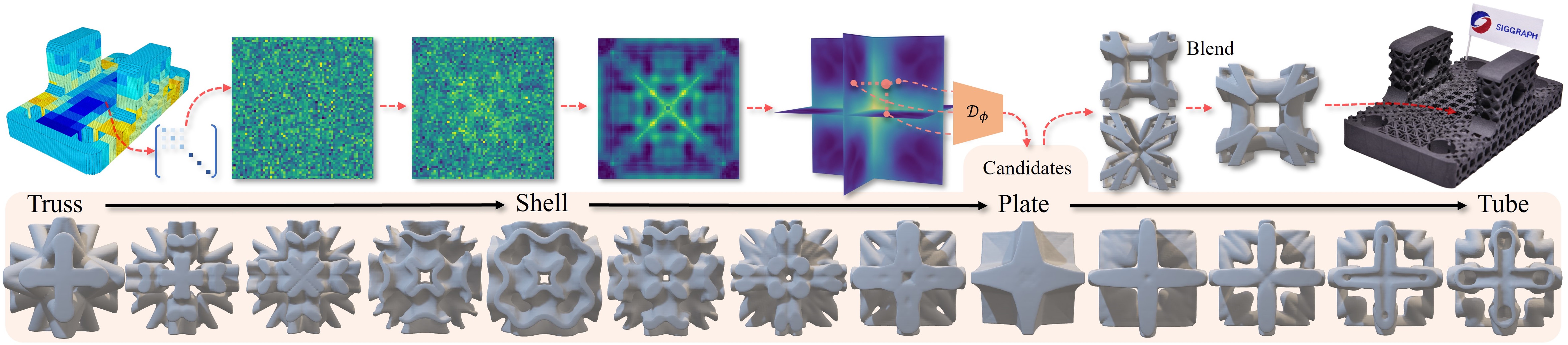}
\caption{
\rev{Overview of \textbf{MIND}, the Microstructure INverse Design system, for generating 3D tileable microstructures with target properties.
A decoder $\mathcal{D}_\phi$ maps latent codes to candidate geometries, enabling diverse and heterogeneous microstructure types, including truss, shell, plate, and tube.}
}
  \label{fig:teaser}
\end{teaserfigure}

\maketitle

\setlength{\belowcaptionskip}{4pt}

\section{Introduction}
\label{sec:introduction}

Metamaterials, often realized as periodic microstructures, have attracted significant attention in both academia and industry, driven by advances in additive manufacturing (AM)~\cite{Kadic2019,Askari2020}. The precise manipulation of microstructures enables control over unique physical properties, such as lightweightness~\cite{lu2014Buildtolast}, elasticity~\cite{schumacher2015Microstructures}, shock protection~\cite{Huang2024}, heat transfer~\cite{Ding2021}, and optical behavior~\cite{yu2014flat}.
Topologies and morphologies govern the physical properties of microstructures, driving extensive research into forward design approaches. These efforts have yielded structures such as struts~\cite{panetta2015Elastic,panetta2017Worstcase}, shells~\cite{liu2022Parametric,Xu2023}, plates~\cite{Tancogne‐Dejean2018}, and stochastic forms~\cite{martinez2016Procedural}. However, forward design, relying on mimetic observations or heuristic rules, cannot exhaustively explore the combinatorial configuration space due to its NP-hard nature~\cite{torquato2005Microstructure, Gao2018}, limiting its ability to achieve the desired macroscopic design targets.

Inverse design, a critical alternative to traditional methods, begins with desired functional properties and works backward to determine the optimal geometry and material distribution. The inverse design of microstructures has been explored for decades, with significant contributions from topology optimization (TO) methods, and has gained momentum through data-driven approaches in recent years~\cite{Lee2023}. However, achieving the goal of generating structures with \emph{precise target physical properties}, \emph{valid geometries}, and \emph{flexibility} for heterogeneous tiling remains an open challenge.
Despite advancements in deep generative neural networks for 3D content creation~\cite{Zhang2024Clay,Wu2024,li2024craftsman,Xu2023Survey}, realizing these objectives remains highly challenging.
The complexity arises from the need for precise control over both geometry and material properties while ensuring their alignment. Maintaining this alignment is particularly difficult, as small changes in geometry can substantially affect physical properties, and vice versa. Furthermore, the relationship between geometry and properties is highly sensitive, requiring careful management in generative models. These issues are further compounded by the ill-posed nature of the inverse problem, where multiple topologies can yield identical or similar effective properties~\cite{Wang2024}.

Most existing approaches compromise by relying on parametric geometric representations, constructing structure-property datasets based on predefined parameters and solving the inverse design problem within specific structure families through inverse searches~\cite{schumacher2015Microstructures, panetta2015Elastic}, differential parameter optimization~\cite{tozoni2020Lowparametric}, or data-driven regression~\cite{bastek2022Inverting, wang2022IHGAN, Zheng2023}. While effective within these classes, they limit design flexibility and structural diversity. A recent approach using self-conditioned diffusion models for microstructure design~\cite{Yang2024} eliminates this restriction, achieving relatively high matching accuracy, but still struggles with maintaining structural integrity.

\rev{In this work, we tackle the inverse design of microstructures with an emphasis on generating architectures that exhibit cubic symmetry.}
Our goal is to achieve a high matching ratio with the desired properties while ensuring geometric validity, including connectivity, cubic symmetry, and boundary compatibility. Furthermore, we aim to enable the generation of diverse microstructure types and morphologies that meet these target properties, without being constrained by predefined parametric classes.

Microstructures differ from general 3D shapes in two key ways: 1) they exhibit high geometric symmetry; and 2) their effective properties are intrinsically tied to their geometry. In fact, discussing a microstructure without considering its properties is meaningless, as the two are inherently coupled.

\rev{
Building on this insight, in contrast to conventional Neural Field Diffusion (NFD)~\cite{Shue2023} which employs triplane representations, we propose a novel hybrid neural representation, \emph{Holoplane}, which embeds \emph{geometric symmetry constraints explicitly} and the \emph{elasticity tensor implicitly}. This combination ensures precise alignment between geometry and properties, resulting in more robust and valid microstructure designs.
}
Additionally, we introduce a latent diffusion-based generative model capable of producing multiple microstructure candidates that satisfy target properties while maintaining connectivity, boundary periodicity, and structural compatibility.

To train and validate our model, we introduce a diverse multi-class dataset encompassing a broad spectrum of geometric morphologies and topologies, including parametric families such as truss, shell, tube, and plate microstructures. We evaluate the system’s performance through the generation of novel microstructures, cross-class interpolation, and the infilling of heterogeneous microstructures. Experimental results demonstrate the model’s ability to generate microstructures that meet target properties, ensure geometric validity, and integrate seamlessly into complex assemblies (\figref{fig:teaser}).

Our main contributions are as follows:
\begin{itemize}[nosep]
\item We tackle the inverse design problem for non-parametric microstructures, enabling the generation of diverse types and morphologies that satisfy target properties through a latent diffusion model.

\item We introduce Holoplane, a hybrid neural representation for microstructures that enhances the alignment between geometry and property distributions, leading to higher property matching accuracy and enhanced validity control over the generated microstructures compared to existing baselines.

\item The proposed latent diffusion framework, built upon Holoplane, also enables optimization of boundary compatibility, achieving superior boundary integration in heterogeneous, multi-scale designs.

\end{itemize}

\section{Related Work}
\label{sec:related}

\subsection{Microstructure Modeling}

The study of forward modeling and optimization of microstructures for metamaterial design has been a prominent area of research since the late 20th century~\cite{Lakes1987}.
A common approach involves representing microstructures using parametric building blocks, with key design variables such as rod radius or shell thickness. This strategy simplifies the design space and facilitates the creation of various morphological types, including truss-based~\cite{nazir2019Stateoftheart, ling2019Mechanical, choi2016Nonlinear, panetta2015Elastic,Liu2022}, shell-based~\cite{bonatti2019Mechanical, overvelde2016Threedimensional, ion2016Metamaterial, ion2018Metamaterial, liu2022Parametric}, and plate-based structures~\cite{wang2020Quasiperiodica,Sun2023}.
Alternatively, microstructures can be designed using mathematical functions, such as Voronoi tessellations and their variants~\cite{martinez2016Procedural, martinez2017Orthotropic, martinez2018Polyhedral}, Gaussian kernels~\cite{tian2020Organic,Bastek2023}, signed distance fields with pre-computed structures~\cite{schumacher2015Microstructures}, or triply periodic minimal surfaces (TPMS)~\cite{hu2020Cellular, yan2020Strong, hu2019Lightweight}.
Although these methods can produce valid geometries, they are limited to specific types of microstructures, creating discrete families within the property space. Interpolating between these families is also challenging.

TO is a key approach for microstructure design~\cite{Sigmund1994, Coelho2007, Zhang2023Optimized}. It aims to identify optimal topologies that meet material property requirements while minimizing cost. By optimizing the density distribution of periodic microstructures and using homogenization to compute effective properties, TO can yield desirable results. However, aside from computational complexity, it faces challenges with boundary compatibility control, particularly when adapting to heterogeneous infilling scenarios~\cite{Cheng2017,wu2018Infill}.

\subsection{Microstructure Inverse Design}

Existing inverse design methods in machine learning and deep learning (ML/DL) can be broadly categorized into three types~\cite{Lee2023}: direct mapping, cascaded neural networks, and conditional generative models.
Direct mapping approaches use regression or surrogate models to directly link material properties to design parameters \cite{Li2020, wang2020DataDriven, bostanabad2019Globally, bastek2022Inverting}. While effective in some cases, these models face limitations such as restricted design spaces and challenges with non-uniqueness and structural similarity.
Cascaded neural networks tackle these issues by combining inverse design and forward modeling in a two-phase approach, ensuring intermediate designs remain consistent with existing data~\cite{Liu2018}.
Conditional generative models, including generative adversarial networks (GANs) \cite{goodfellow2020Generative} and variational autoencoders (VAEs) \cite{KingmaVAE}, are widely used for achieving one-to-many mappings. Previous attempts \cite{zheng2021Controllable, wang2022IHGAN}, \cite{Ma2019, wang2020Deep} have focused on preset structure classes. More recent work has explored denoising diffusion probabilistic models (DDPM) \cite{Ho2020DDPM} for the inverse design of 2D \cite{Wang2024} and 3D \cite{Yang2024} microstructures. However, these models struggle with matching geometry and property distributions, particularly when handling diverse or class-free microstructure data, which reduces their matching accuracy.

From the perspective of the representation of microstructures, existing inverse design methods can be generally categorized into two main types:
(1) Pixel or voxel-based representations, which have been effective for modeling composite materials and multi-phase structures that occupy the entire microstructure volume~\cite{li2018Deep, Li2020, Hsu2020, noguchi2021Stochastic}. However, this approach is highly inefficient for single-material structures or sparse 3D geometries~\cite{Yang2024}, and neglects the geometric connectivity constraints, which are crucial for microstructures fabricated from a single material.
(2) Pre-computed parametric representations, which utilize specific families of design parameters of building blocks~\cite{wang2021DataDriven,Zheng2023,Ha2023}, implicit functions  (e.g., TPMS \cite{Li2019} or spinodoids \cite{Kumar2020}), and are typically solved through inverse searches~\cite{schumacher2015Microstructures, panetta2015Elastic}, differential parameter optimization~\cite{tozoni2020Lowparametric}, or data-driven regression~\cite{bastek2022Inverting, zheng2021Controllable, wang2022IHGAN}. Although these methods offer greater computational efficiency, they are constrained by predefined structural classes, which restrict the diversity and flexibility of the design space.

In contrast, implicit neural representations~\cite{Deng2024neural, Shue2023, chan2022triplane} effectively enhance geometric integrity while maintaining resolution independence in the generated geometries.
Recent work on metamaterial sequences within plate lattices \cite{Zhao2024} has demonstrated the effectiveness of implicit neural representations. Building on this, our approach further integrates explicit encoding to enforce microstructural symmetry.

Recent studies have also explored mapping nonlinear properties to 2D microstructure designs using diffusion models, going beyond linear elastic properties~\cite{Bastek2023, Vlassis2023, Li2023, Park2024}.
These approaches share similar strategies, aiming to limit the degrees of freedom in the structures and reduce the complexity of the property distribution space, thereby enhancing matching accuracy during training. However, challenges persist in aligning non-parametric geometries with their corresponding properties.

\section{Overview}
\label{sec:overview}

\subsection{Problem Statement}
We propose a framework for generating tileable microstructures with targeted physical properties.
The geometry of each microstructure $\Omega$ is represented by a signed distance field (SDF) $\phi_{\Omega}(\mathbf{x}): \mathbb{R}^3 \to \mathbb{R}$.
By generating the structure to be translationally symmetric, we enable tessellation in 3D space, ensuring the structure remains invariant under lattice translations:
\begin{equation}
    \phi(\mathbf{x})=\phi(\mathbf{x} + n\mathbf{t}),
\end{equation}
where $\mathbf{t}$ is a lattice translation vector and $n$ is the tiling number.

\rev{The mechanical properties of the microstructure are described by the macroscopic elasticity tensor $\mathbf{C} \in \mathbb{R}^{6\times6}$, which can be directly converted to Young's modulus $E$, Poisson's ratio $\nu$, and shear modulus $G$ (\srefhomo).}
Our objective is to generate microstructures whose elasticity tensor $\mathbf{C}(\phi_{\Omega}(\mathbf{x}))$ closely matches a target elasticity tensor $\mathbf{C}_{\text{target}}$.

\subsection{MIND: Neural Microstructure Generation}
\begin{figure*}[ht]
    \centering
    \includegraphics[width=\linewidth]{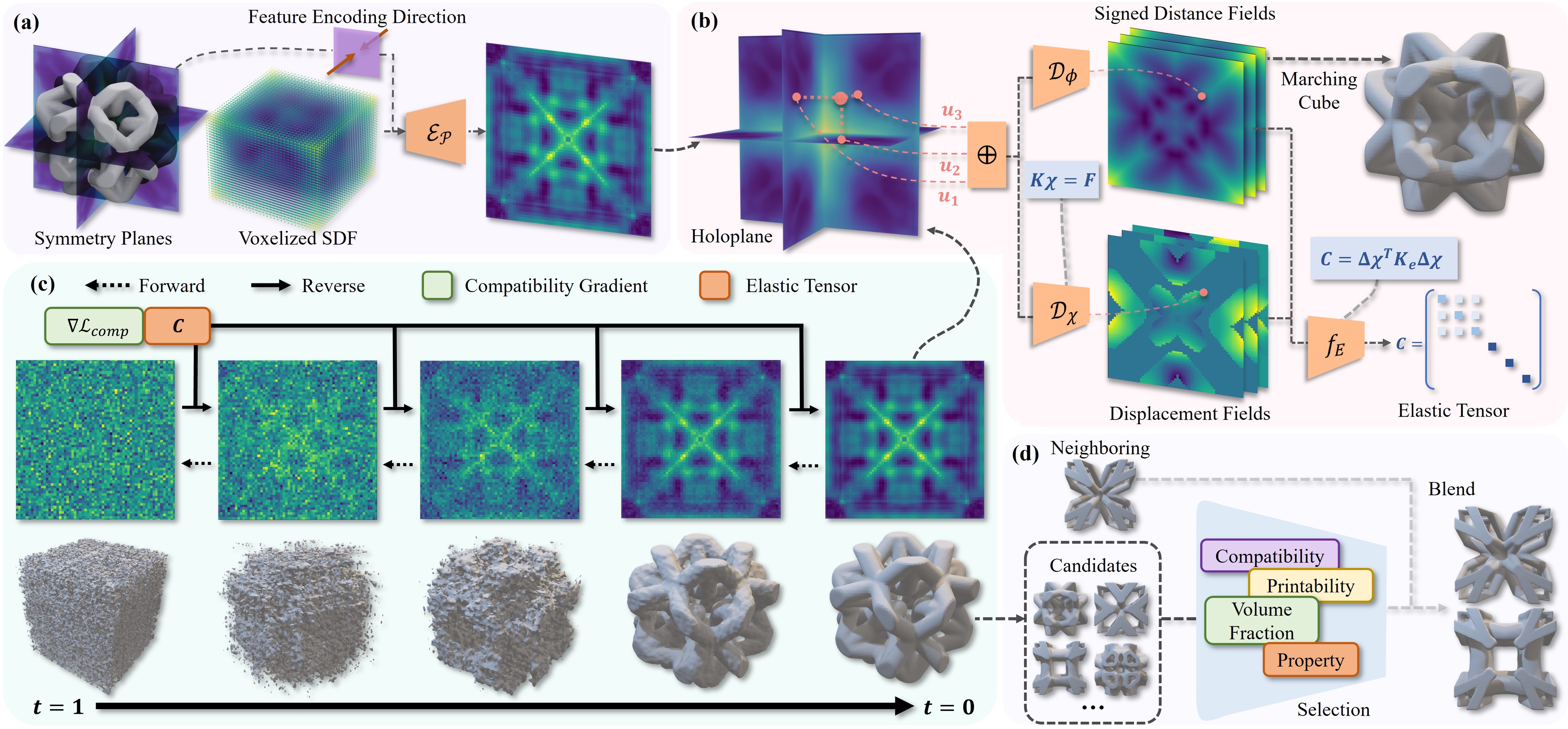}

    \caption{
    \rev{
    Pipeline of \textbf{MIND}.
    (a) We explicitly encode the microstructural symmetry in voxel space with an encoder $\mathcal{E}_\holo$, leveraging the inherent symmetry to ensure tilability.
    (b) A combination of SDFs and displacement fields integrates physical priors into the implicit fields, enabling a hybrid neural representation, \emph{Holoplane}.
    Neural decoders $\mathcal{D}_\phi$ and $\mathcal{D}_\chi$ generate SDFs and displacement fields, respectively.
    (c) Holoplane can be conditionally generated using a diffusion model, yielding diverse microstructural demands.
    (d) We apply this process to heterogeneous design, generating seamlessly fitting, 3D-printable structures.
    }
    }

    \label{fig:pipeline}
\end{figure*}

To achieve this, we introduce a latent representation, termed \textit{Holoplane}, and train an autoencoder to encode microstructures into this latent space (Fig.~\ref{fig:pipeline}).
Diffusion models are then employed for conditional generation within the latent space.

\begin{wrapfigure}{l}{0.4\linewidth}
    \centering
    \setlength{\columnsep}{5pt}
    \setlength{\intextsep}{5pt}
    \includegraphics[width=1.0\linewidth]{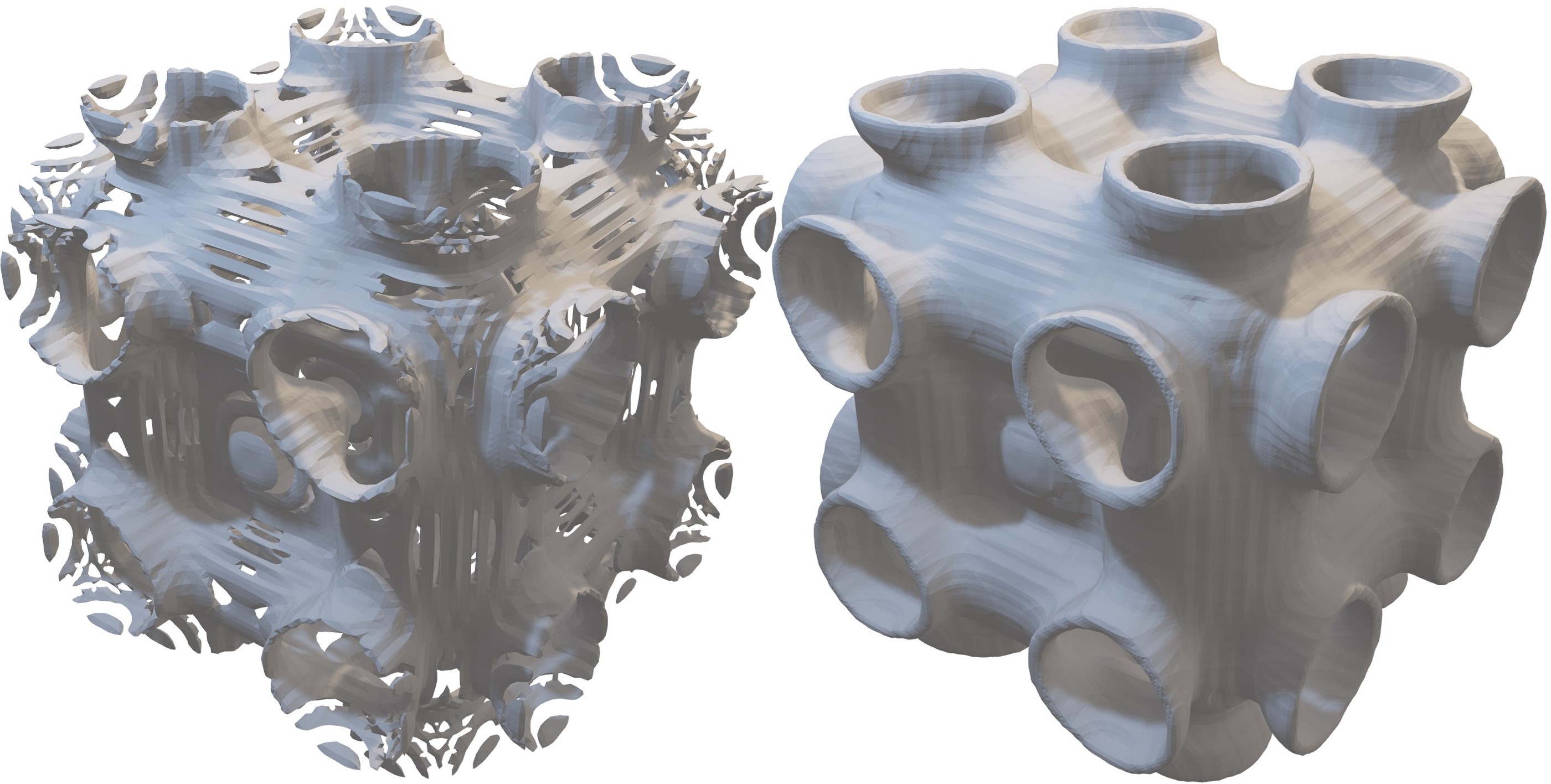}
    \caption{\rev{Structures at resolution 128 (left) and super-resolution 196 (right), generated from the SDF zero isosurface via marching cubes.}
    }
	\label{fig:super_resolution}
\end{wrapfigure}

Tileable microstructures often possess inherent symmetries, which can be explicitly utilized to achieve a more compact and efficient representation.
Voxel-based representations are particularly well-suited for encoding such structural symmetries.
For example, previous work~\cite{Yang2024} has utilized the $\frac{1}{8}$-space of the microstructure to express tetrahedral symmetry.
However, voxel-based encodings are inherently limited by their resolution.
As illustrated in Fig.~\ref{fig:super_resolution}, even at a relatively high resolution of $128^3$, typical structures cannot be faithfully represented.

\rev{
To address this, we introduce a hybrid explicit-implicit representation method for encoding microstructures.
In this context, 'hybrid' specifically refers to the combination of explicit voxel grids, where geometric symmetries are directly enforced, and implicit neural field-based representations, such as Signed Distance Functions (SDFs) and displacement fields, which enable continuous structure representation and precise geometry-property alignment (Fig.~\ref{fig:pipeline} (a)).
This integration ensures accurate symmetry capture and robust microstructure modeling.
}
We refer to this representation as \textit{Hybrid Symmetric Representation}~(Sec.~\ref{sec:sym_enc}).

The geometry of a microstructure strongly influences its stiffness, but the relationship is highly
 nonlinear.
Minor geometric changes can cause substantial variations in properties.
Solely encoding geometry risks the autoencoder learning spurious correlations, hindering diffusion-based generation in the latent space.
To overcome this, we incorporate physical priors during the training of the autoencoder, enabling the model to jointly capture both geometric and physical details (Fig.~\ref{fig:pipeline} (b)).
This approach is termed \textit{Physics-aware Neural Embedding}~(Sec.~\ref{sec:phy_enc}).

By combining these ideas, we present a novel representation, termed \textit{Holoplane}.
The Holoplane $\mathcal{P} \in \mathbb{R}^{r\times r\times c}$ can be viewed as a symmetric 2D snapshot of the microstructure’s geometry $\phi$ and physical properties $\mathbf{C}$, aligning them within a unified latent space. Here, $r$ represents the resolution of the snapshot, while $c$ denotes the number of channels.

We train a diffusion model to generate Holoplanes conditioned on given properties (Sec.~\ref{sec:diffusion}).
The generated Holoplanes are then decoded to produce the microstructure SDF $\phi_{\Omega}$.
When performing heterogeneous design, the compatibility between adjacent microstructures significantly impacts the overall physical performance.
To address this, we utilize the gradient of a boundary-compatibility loss to guide the diffusion sampling of Holoplanes, ensuring that the generated microstructures adhere to compatibility constraints.
Additionally, we apply interpolation-based blending to ensure seamless alignment of microstructure boundaries, achieving a tight and consistent fit at their interfaces.

\section{Representing Structures with Holoplanes}
\label{sec:holoplane}
The Holoplane representation consists of feature maps that encode both the geometric shape and physical properties of the microstructure within a symmetry-preserving, physics-aware autoencoder.

\rev{To obtain the holoplane representation $\holo$, we project $\phi_\Omega(\bx)$ onto a set of planes that corresponding to the cubic symmetry. }
The projection is defined as:
\begin{equation}
    \holo_k=\mathcal{E}(\phi_\Omega;\theta,k),
\end{equation}
where $\mathcal{E}$ is a neural encoder that maps the SDF onto the $k$-th symmetry plane of microstructure.
\rev{For symmetry groups other than cubic, the number of planes may be modified correspondingly (\sreftraining). }

During decoding, the Holoplane representation is used to reconstruct a neural field $\mathcal{D}(\bx)$ at any given point.
The coordinates $\bx$ are first projected onto the symmetry planes, and the corresponding features are then sampled and processed by a decoder:
\begin{equation}
\label{eq:decode}
    \mathcal{D}(\holo, \bx) = f_{\text{decode}}(\holo_1(\bu_1(\bx)) + ... + \holo_k(\bu_k(\bx)); \theta),
\end{equation}
where $\bu(\bx)$ represents the coordinate projected onto the plane, $\holo(\bu)$ represents the sample from the plane, and $f_\text{decode}$ is a lightweight multilayer perceptron (MLP) that combines the features from all the planes.

\subsection{\rev{Symmetry-aware Representation}}
\label{sec:sym_enc}
\rev{A symmetry plane $\bu$ is defined as one where the SDF values of points reflected across the plane are identical.}
The SDF is discretized as voxel data and fed into the encoder $\mathcal{E}$.
The decoder generates an SDF $\mathcal{D}\phi$, which is compared against the ground truth through the reconstruction loss:
\begin{equation}
\label{eq:rec}
    \mathcal{L}_{\phi}=\sum_{\bx}{\left|\left|\phi_\Omega(\bx) - \mathcal{D}_\phi(\holo_{\Omega}, \bx)\right|\right|}^2.
\end{equation}
The loss is minimized by sampling random points within the SDF (Sec.~\ref{sec:network_imp}), ensuring accurate reconstruction of the implicit field.

\subsection{Physics-aware Neural Embedding}
\label{sec:phy_enc}

We enhance the Holoplane's ability to align with physical properties by integrating physical priors.
Mapping diverse families of microstructure data to their corresponding physical properties poses a significant challenge, as a simple MLP~\cite{Zheng2023} tends to overfit on our dataset.
To address this, we integrate physical equations derived from the homogenization process~\cite{andreassen2014Design, dong2018149}, enabling the construction of a smooth and interpretable latent space.

\rev{
The latent space is constrained by minimizing a property loss between the ground-truth properties $\bC_\Omega$ and the output of a solver $f_{E}$, which computes the elasticity tensor from the Holoplane representation differentiably.
We leverage the displacement fields $\boldsymbol{\chi}$, introduced during the homogenization process (\srefhomo), to bridge geometry and physical properties in a more coherent and efficient manner.
The calculation of $\boldsymbol{\chi}$ is formulated as solving the system $\bK \boldsymbol{\chi} = \bF$, where $\mathbf{K}$ is the global stiffness matrix and $\mathbf{F}$ represents the applied load.
This step is represented using:
}
\begin{equation}
\label{eq:disp}
    \mathcal{L}_{\boldsymbol{\chi}}=\sum_{\bx}{\left|\left|{\boldsymbol{\chi}_{\Omega}}(\bx) - \mathcal{D}_{\boldsymbol{\chi}}(\holo_{\Omega}, \bx)\right|\right|}^2.
\end{equation}

\begin{figure}[hbt]
    \includegraphics[width=\linewidth]{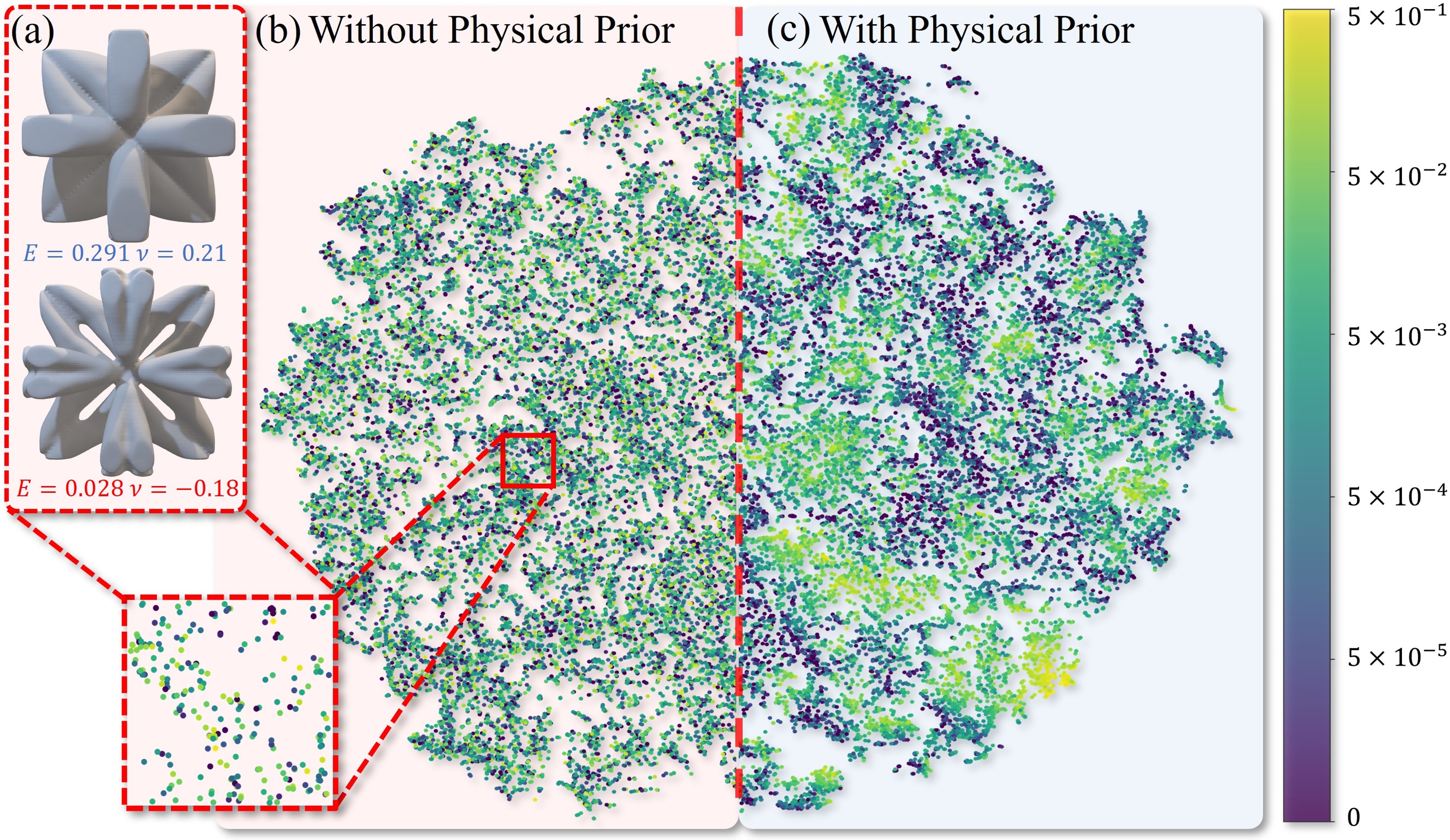}
    \caption{We visualize the latent space with and without physical priors using t-SNE. The color represents Young's moduli.
    Without physical priors, close data points exhibit significant property discrepancies (a).
    This disordered latent space (b) hinders the Diffusion model's ability to condition effectively.
    Incorporating physical priors improves this distribution significantly (c).
    We further conducted ablation experiments to compare the results of using the generative model in space (b) and (c) (Sec.\ref{sec:accuracy}).}
	\label{fig:distribution}
\end{figure}

We further leverage these two fields $\phi$ and $\boldsymbol{\chi}$ to predict the elasticity tensor by minimizing the property loss:
\begin{equation}
    \mathcal{L}_{E}={||\bC_\Omega - \sum_{\bx \in \Omega}f_{E}(\mathcal{D}_{\boldsymbol{\chi}}(\bx), \mathcal{D}_\phi(\bx); \holo_\Omega)||}^2.
\end{equation}
\rev{
Through this physics-aware embedding, displacement fields $\boldsymbol{\chi}$ effectively introduce physical information into the latent space by guiding the optimization process via gradient-based backpropagation (Fig.~\ref{fig:distribution}).
}

\section{Generating Structures with Diffusion}
\label{sec:diffusion}

We define a representation of microstructures through Holoplanes, where each Holoplane serves as an encoding of the microstructure, and decoding reconstructs both the structure and its physical properties.
A conditional diffusion model is trained to generate diverse Holoplanes, enabling the design of microstructures with specific properties.

Each Holoplane instance represents a sample from the distribution $\holo \sim p(\holo)$.
During the training phase, the diffusion process perturbs the original distribution by adding Gaussian noise $\epsilon \sim \mathcal{N}(0, \sigma \mathbf{I})$ to the original data~\cite{Ho2020DDPM}.
A neural network is trained to learn the denoising process.

Following~\cite{song2021scorebased, karras2022edm}, the diffusion process is defined as:
\begin{equation}
\label{eq:dif_sample}
    d\holo=-\dot{\sigma}(t)\sigma(t)\nabla_{\holo}\log{p(\holo;\sigma(t))}dt,
\end{equation}
where $\sigma(t)$ is the noise level at time $t \in (0, 1]$.

A neural denoiser $\Psi(\holo;\sigma, \theta)$ is trained by minimizing:
\begin{equation}
\label{eq:dif_train}
    \mathbb{E}_{\holo\sim p(\holo)}\mathbb{E}_{\mathbf{\epsilon \sim \mathcal{N}(0, \sigma \mathbf{I})}} \left|\left| \Psi(\holo + \epsilon; \sigma) - \holo\right|\right|_2^2.
\end{equation}
This denoising process can be interpreted as learning to reverse the noise perturbation, recovering the original microstructure from noisy samples.
The gradient of the log-probability in Eq.~\ref{eq:dif_sample} is then expressed as $\nabla_{\holo}\log{p(\holo;\sigma)}=\left(\Psi(\holo; \sigma) - \holo\right) / \sigma^2$.

\subsection{Guided Generation Using Properties}
A conditional denoiser $\Psi_C$ is trained according to Eq.\ref{eq:dif_train} with $p_{data}=p(\holo | C)$, where $C$ represents the elastic tensor of microstructures.
To improve properties-conditioned generation, we incorporate Classifier-free Guidance (CFG ~\cite{ho2022classifierfree}).
An unconditional denoiser $\Psi_{\mathbf{0}}$ is trained without properties.
During inference, CFG interpolates between the conditional and unconditional outputs to guide the sampling process:
\begin{equation}
    \Psi(\holo;\sigma, C)=\Psi_{\mathbf{0}} + w(\Psi_C - \Psi_{\mathbf{0}}),
\end{equation}
where $w$ controls the strength of the guidance.

\subsection{Boundary Compatibility Enhancement}
\label{sec:compat}

\paragraph{Compatibility gradient}
In heterogeneous design, ensuring continuous boundaries between adjacent cells is crucial.
To achieve this, we introduce a compatibility gradient that enforces consistent boundary shapes between two microstructures during diffusion sampling.
The compatibility loss $\mathcal{L}_{\text{compat}}$ is designed to minimize the discrepancy between the boundaries of two microstructures.
By using the Holoplane, we can more effectively compare and align the microstructures within a shared latent space.
Therefore, the compatibility loss is formulated as:
\begin{equation}
    \mathcal{L}_{\text{compat}} = \int_{\Gamma} \| \holo_A - \holo_B \|^2 \, d\mathbf{x},
\end{equation}
where $\Gamma$ denotes the boundary area.
To maintain consistent boundary shapes throughout the sampling process, we modify the ODE (Eq.~\ref{eq:dif_sample}) as follows:
\begin{equation}
    d\holo=-\dot{\sigma}(t)\sigma(t)(\nabla_{\holo}\log{p(\holo;\sigma(t))} - \nabla_{\holo}\mathcal{L}_{\text{compat}})dt.
\end{equation}

\begin{figure}[tb]
    \includegraphics[width=\linewidth]{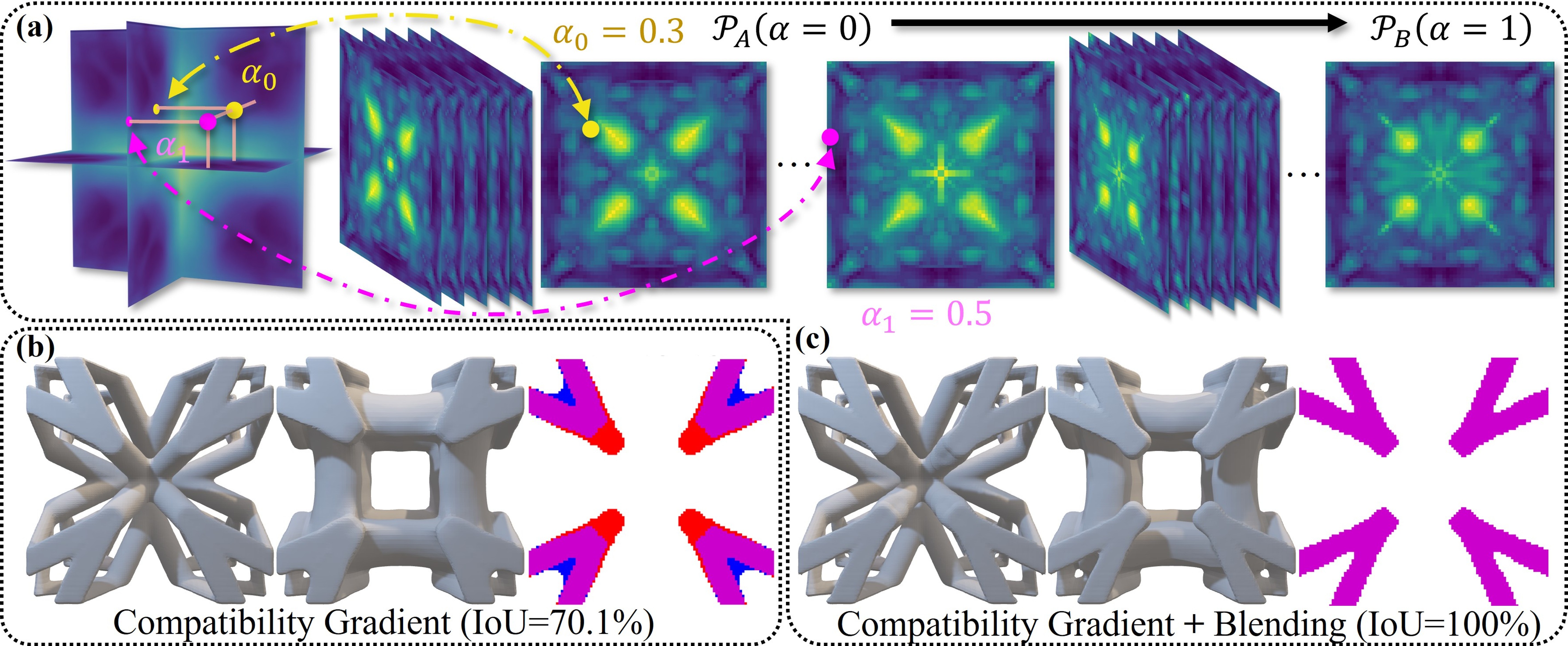}
    \caption{We enhance the boundary compatibility through smooth interpolation. (a) The blended microstructure’s SDF is decoded from the corresponding positions in the interpolation sequence.
    We quantify boundary similarity using the intersection-over-union (IoU) of binarized boundary surfaces.
    (b) Using only the boundary compatibility gradient, we achieve 70.1\% boundary similarity.
    (c) Under blending, boundary compatibility reaches 100\%.}
	\label{fig:blending}
\end{figure}

\paragraph{Blending}
Although the compatibility gradient promotes alignment along the boundaries, minor discontinuities may still arise (Fig.~\ref{fig:blending} (b)).
To further enhance compatibility, we adopt an interpolation-based blending approach.

First, we add noise to the two Holoplanes during the forward diffusion process, resulting in noisy representations $\holo^{\sigma}$.
Next, spherical linear interpolation (slerp) is applied between these two noisy data points using a coefficient $\alpha$:
\begin{equation}
    \holo_\alpha^{\sigma}=\text{slerp}(\holo^{\sigma}_A, \holo^{\sigma}_B; \alpha).
\end{equation}
We then perform reverse diffusion to generate an interpolated Holoplane, $\holo_{\alpha}$.

The interpolated results are then used to reconstruct the boundary region, seamlessly blending the microstructures (Fig.~\ref{fig:blending}(a)):
\begin{equation}
    \phi(\bx) = \mathcal{D}_{\phi}(\holo_{\alpha}, \bx).
\end{equation}
Specifically, for a coordinate $\bx$ located at a distance $x_0$ from the boundary of microstructure $A$ (adjacent to $B$), we compute an interpolation coefficient $\alpha = \frac{l - x_0}{2l}$, where $l$ represents the boundary width.
Interpolation-based blending ensures perfect connectivity (Fig.~\ref{fig:blending} (c)) at the boundaries, enhancing structural stability during heterogeneous design.

\section{Implementation}
\label{sec:implementation}

\subsection{Dataset}
\label{sec:dataset}
\begin{figure}[!]
    \centering
    \includegraphics[width=0.9\linewidth]{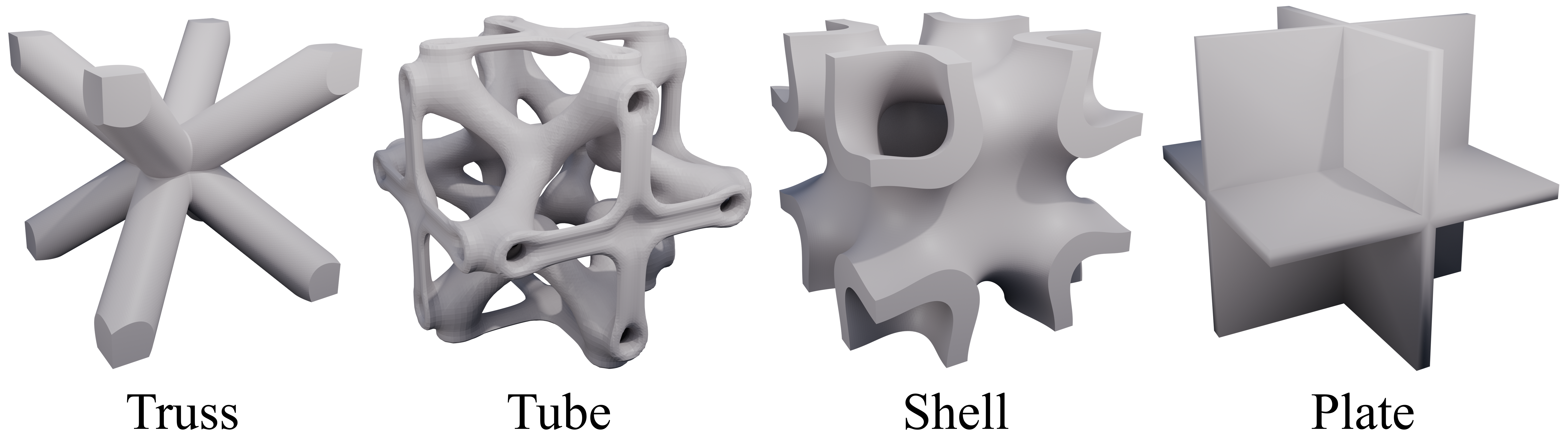}
    \caption{Exemplar models for four different structure types.}
    \label{fig:dataset}
\end{figure}

In this work, we focus on microstructures with cubic symmetry $O_h$, a high-order symmetry group that satisfies translational symmetry.
Our dataset consists of a mixed set of microstructure meshes, including truss (33\%), tube (38\%), shell (17\%), and plate (12\%) structures (Fig.~\ref{fig:dataset}), with a total of 180,000 samples and a volume fraction ranging from 5\% to 65\%.
The skeletons of the truss and tube structures were generated using the method proposed by~\cite{panetta2015Elastic}, while the shell and plate structures were constructed following the parametric methods introduced by~\cite{liu2022Parametric} and~\cite{Sun2023}, respectively.
We randomly selected 1,000 samples as the test set and validation set respectively.

\subsection{Properties Calculation}
\label{sec:properties}
We calculate the elastic tensor $\mathbf{C}$ for all data using a GPU-accelerated implementation of the homogenization method~\cite{andreassen2014Design, dong2018149}.
In the calculations, the base material's Young’s modulus $E$ is set to 1 (dimensionless), and the Poisson’s ratio $\nu$ is set to 0.35.
Under $O_h$, the elastic properties of a microstructure reduce to three independent components: $C_{11}, C_{12}, C_{44} \in \mathbf{C}$.

\subsection{Network and Training}
\label{sec:network_imp}
\rev{Our decoder performs downsampling along the normal direction of the symmetry plane.
In experiments with cubic symmetry, it is sufficient to choose one of the three equivalent axial planes as the symmetry plane.}
The network compresses the SDF $\phi \in \mathbb{R}^{128\times128\times128\times1}$ into a 2D latent Holoplane $\holo \in \mathbb{R}^{64\times64\times32}$, where 32 denotes the number of channels.
During decoding, after projecting any point $\bx$ onto the Holoplane to obtain $\bu$, we feed $\holo(\bu)$ into a 5-layer residual MLP to predict the corresponding fields.
\rev{
To achieve smoother field representations, we follow the approach in \cite{Shue2023} and introduce total variation (TV) loss and explicit density regularization (EDR) loss.
We also use an $L_2$-norm to regularize the Holoplane distribution toward a standard normal distribution.
The total loss for training the autoencoder is expressed as:
\begin{equation}
    \mathcal{L} = \mathcal{L}_\phi +
                  \lambda_{0}\mathcal{L}_{\chi} +
                  \lambda_{1}\mathcal{L}_{E} +
                  \lambda_{2}\mathcal{L}_{\text{TV}} +
                  \lambda_{3}\mathcal{L}_{\text{EDR}} +
                  \lambda_{4}\mathcal{L}_{2}.
\end{equation}
}

We employ the EDM (Elucidated Diffusion Model,~\cite{karras2022edm}) framework as our generative backbone, where timestep is set to 18.
Due to the high symmetry of $O_h$, we use only one-quarter of the Holoplane.
The guidance strength is set to 7 to balance diversity and adherence to the conditioned properties.
For further details, refer to \sreftraining.

\subsection{Metrics}
\label{sec:metrics}
We use two metrics introduced in~\cite{Yang2024} to evaluate the inverse generation.
\rev{The error in microstructure properties is computed component-wise as follows:
\begin{equation}
\label{eq:error}
    \text{Err}_{ij} = \frac{\left|C_{ij}^{\text{pred}} - C_{ij}^{\text{target}}\right|}{C_{ij}^{\text{max}} - C_{ij}^{\text{min}}}.
\end{equation}
}
For two given microstructures, we first binarize their SDFs and then calculate the similarity using the following formula:
\begin{equation}
\label{eq:sim}
    \text{Sim}(\Omega_1, \Omega_2) = \frac{\sum_{v \in V} \mathbb{V}_{\Omega_1(v) = \Omega_2(v)}}{\sqrt{|\Omega_1| \cdot |\Omega_2|}},
\end{equation}
where the numerator is the number of intersecting voxels between the two structures, and the denominator is the square root of the product of the total voxel counts of both structures.

\subsection{Printable Heterogeneous Design}
Heterogeneous design aims to optimize material properties for specific performance.
We propose a linear finite element method (FEM) to optimize mechanical strength, detailed in \srefHeterdesign.
The algorithm computes $\bC_{\text{target}}$ for each hexahedral cell, which is then used to generate microstructures via MIND.
For each cell, 64 candidates are generated, and the best is selected based on printability, volume fraction, and accuracy.
To ensure printability, microstructures of size $S$ must have a minimum feature size exceeding the printer precision $\epsilon$.
Feature size is evaluated via SDF connectivity across resolutions, with a detection resolution of $R = S / \epsilon$ as the filtering threshold.
Microstructures are generated sequentially, ensuring boundary compatibility (Sec.~\ref{sec:diffusion}) for structural continuity and printability.

\section{Results}

\subsection{Accuracy}
\label{sec:accuracy}
We utilized the properties of the test set (Sec.~\ref{sec:dataset}) as conditions for the inverse generation test.
For each microstructure, 8 candidate models were generated, resulting in a total of 8,000 models.
Tests were performed on a 4xA40 server, with an average generation time of \textbf{0.13 seconds} per microstructure and \textbf{0.02 seconds} for SDF decoding.
Properties were computed using the homogenization method, and errors were calculated according to Eq.\ref{eq:error}.

\begin{figure}[hbt]
    \centering
    \includegraphics[width=\linewidth]{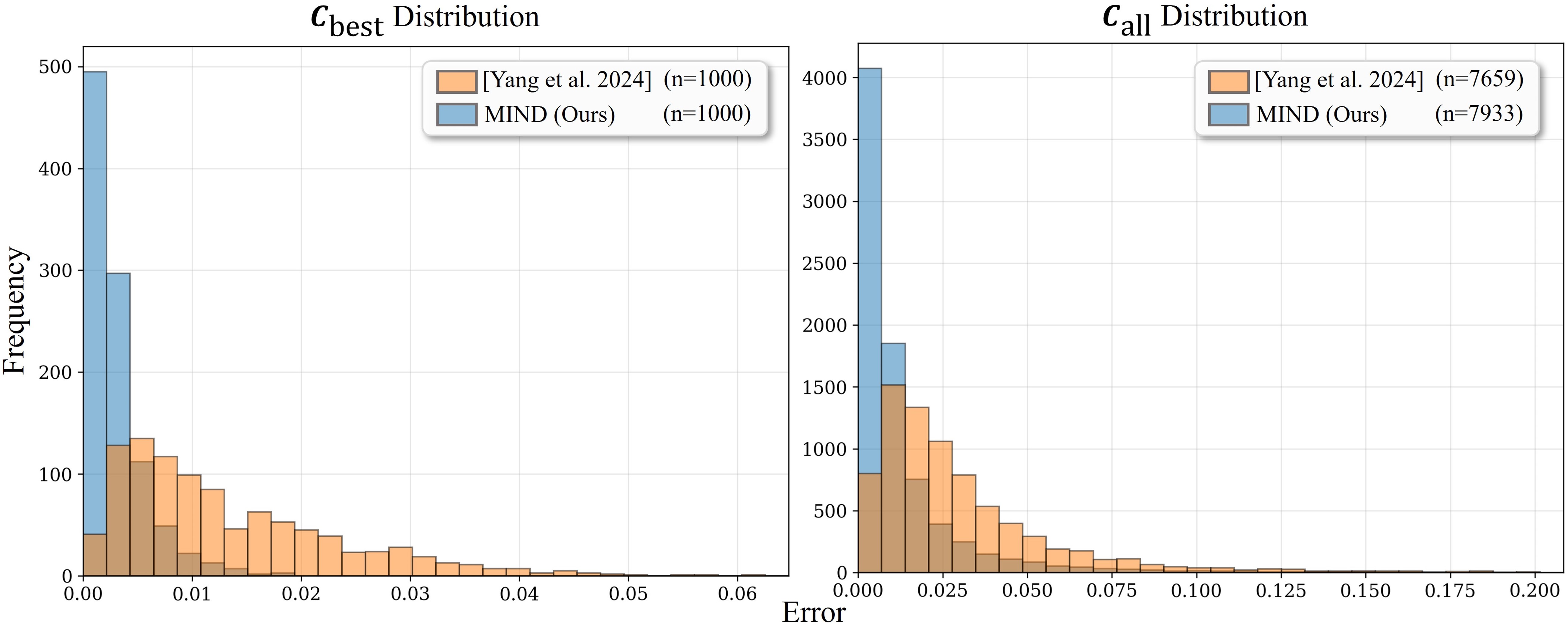}
    \caption{The error distribution of MIND and \cite{Yang2024} on the same test dataset.}
    \label{fig:err_dist}
\end{figure}

Tab. \ref{tab:error_comparison} and Fig.~\ref{fig:err_dist} present a comparison between our method and \cite{Yang2024}, demonstrating that our approach achieves state-of-the-art performance in inverse generation of microstructures.
Moreover, our method can be viewed as a variant of the triplane representation \cite{chan2022triplane, Shue2023}.
To assess the necessity of the Holoplane, we conducted ablation studies using the triplane representation (NFD~\cite{Shue2023}).
Among the generated models, some were excluded due to issues with translational symmetry or connectivity, preventing property calculations.
The physical validity ratio was \textbf{95.7\%} for \cite{Yang2024}, \textbf{97.8\%} for the NFD representation, and \textbf{99.2\%} for our method.
The results strongly emphasize that our Holoplane representation significantly enhances the accuracy of property predictions.

\begin{table}[hb]
    \centering
    \caption{
    The properties of the test set are used as the conditions to evaluate generation errors.
    $C_{\text{best}}$ is the average of the best result from each group of 8, while the other columns represent the average of all 8000 structures. The NFD + Phy approach leverages the physical-aware embedding introduced in Sec.~\ref{sec:phy_enc} to align geometry and physics within the latent space.}
    \label{tab:error_comparison}
    \begin{tabular}{lcc|ccc}
        \toprule
        Method                      & $\bC_{\text{best}}$  & $\bC_{\text{all}}$ & C\textsubscript{11} & C\textsubscript{12} & C\textsubscript{44} \\
        \midrule
        \cite{Yang2024}             & 1.33\% & 2.96\% & 2.50\% & 3.68\% & 2.70\% \\
        NFD                         & 0.63\% & 5.39\% & 5.28\% & 4.86\% & 6.01\% \\
        NFD + Phy                   & 0.44\% & 1.68\% & 1.49\% & 1.80\% & 1.75\% \\
        MIND (Ours)                 & \textbf{0.29\%} &\textbf{ 1.27\% }& \textbf{1.13\%} &\textbf{ 1.33\%} & \textbf{1.34\%} \\
        \bottomrule
    \end{tabular}
\end{table}

\subsection{Generation Boundary}
\begin{figure}
    \centering
    \includegraphics[width=\linewidth]{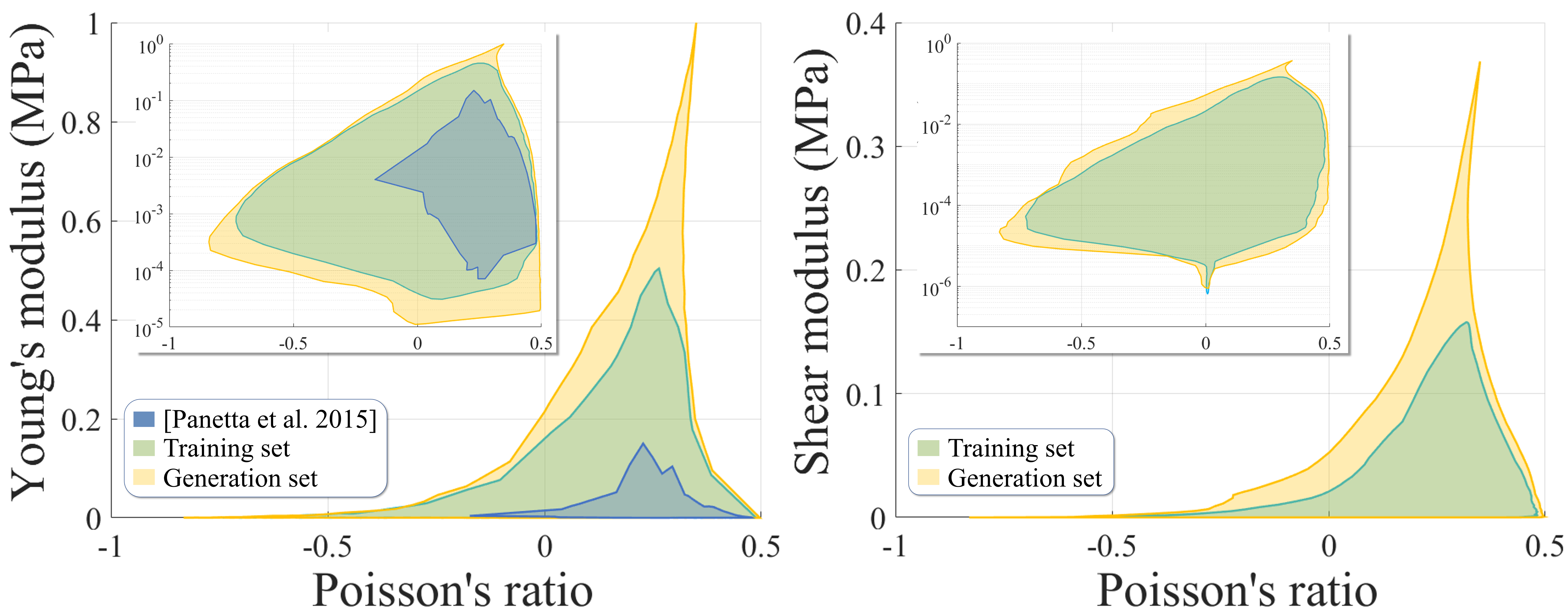}
    \caption{
    \rev{
    Comparison of the property space between the training set and the generation set.
    The training set consists of approximately 180,000 samples, as described in Sec.~\ref{sec:dataset}, while the generation set contains around 550,000 samples obtained by random sampling inside and near the boundary of the property space.
    We visualize both the original distribution and the log-scale distribution of the property space (top-left corner).
    Our method effectively extends the boundary of the property space, significantly increasing the maximum Young's modulus and shear modulus while also achieving a lower negative Poisson's ratio.}}
    \label{fig:design_space}
\end{figure}

To explore the boundary of our network’s generative capacity, we randomly sample points near the boundary of the property space.
This process continues until the network fails to generate structures that meet the specified properties.
As shown in Fig.~\ref{fig:design_space}, our method successfully expands the design space.

\subsection{Diversity}
\begin{figure}
    \centering
    \includegraphics[width=\linewidth]{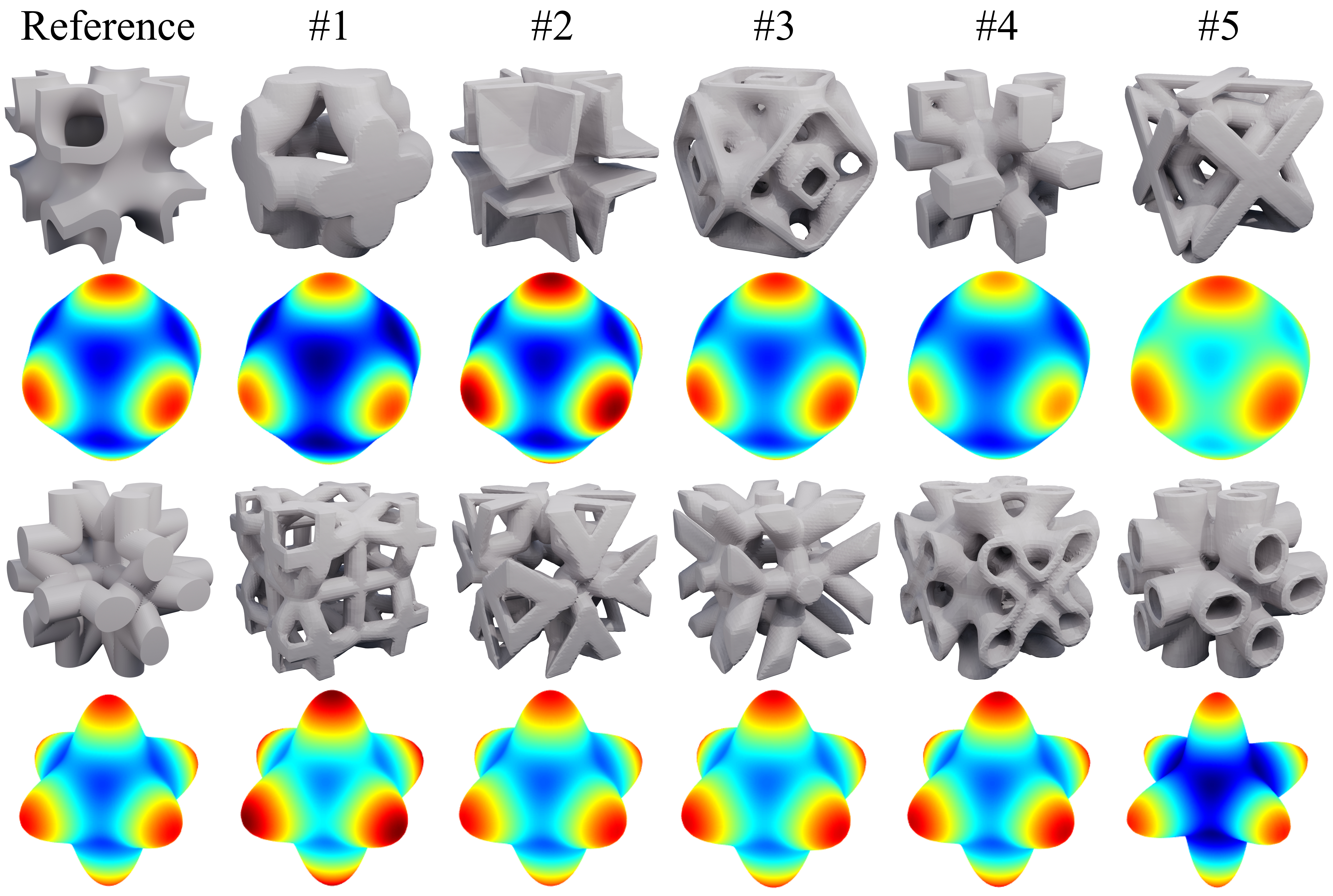}
\caption{
Inverse generation from a reference model.
Using the reference model’s mechanical properties as input, five candidate models generated by our framework are listed.
Each model’s Young’s modulus surface is shown with a consistent color bar, demonstrating structures in diverse morphologies with similar mechanical properties.
\rev{
The Young’s modulus surface represents the directional variation of Young’s modulus on a unit sphere, computed according to the method described by~\cite{liu2021MPP} on mechanical property profiles of microstructures via asymptotic homogenization.
}
}
    \label{fig:shape_gen}
\end{figure}

We compute the average similarity between the 8,000 generated microstructures and the entire training set.
Our model achieves an average similarity of \textbf{81.52\%}, which is lower than the \textbf{93.48\%} reported in \cite{Yang2024}.
Besides, we selected some representative models and utilized their mechanical properties ($E, \nu, G$) as input for inverse generation.
As shown in Fig.~\ref{fig:shape_gen}, given a target property as input, our method can generate multiple distinct types of structures while maintaining similar properties.
This indicates that our model exhibits greater shape diversity and is not merely memorizing the training data.
Additional results are visualized in \srefmorevis.

\subsection{Interpolation}
\begin{figure}
    \centering
    \includegraphics[width=1.0\linewidth]{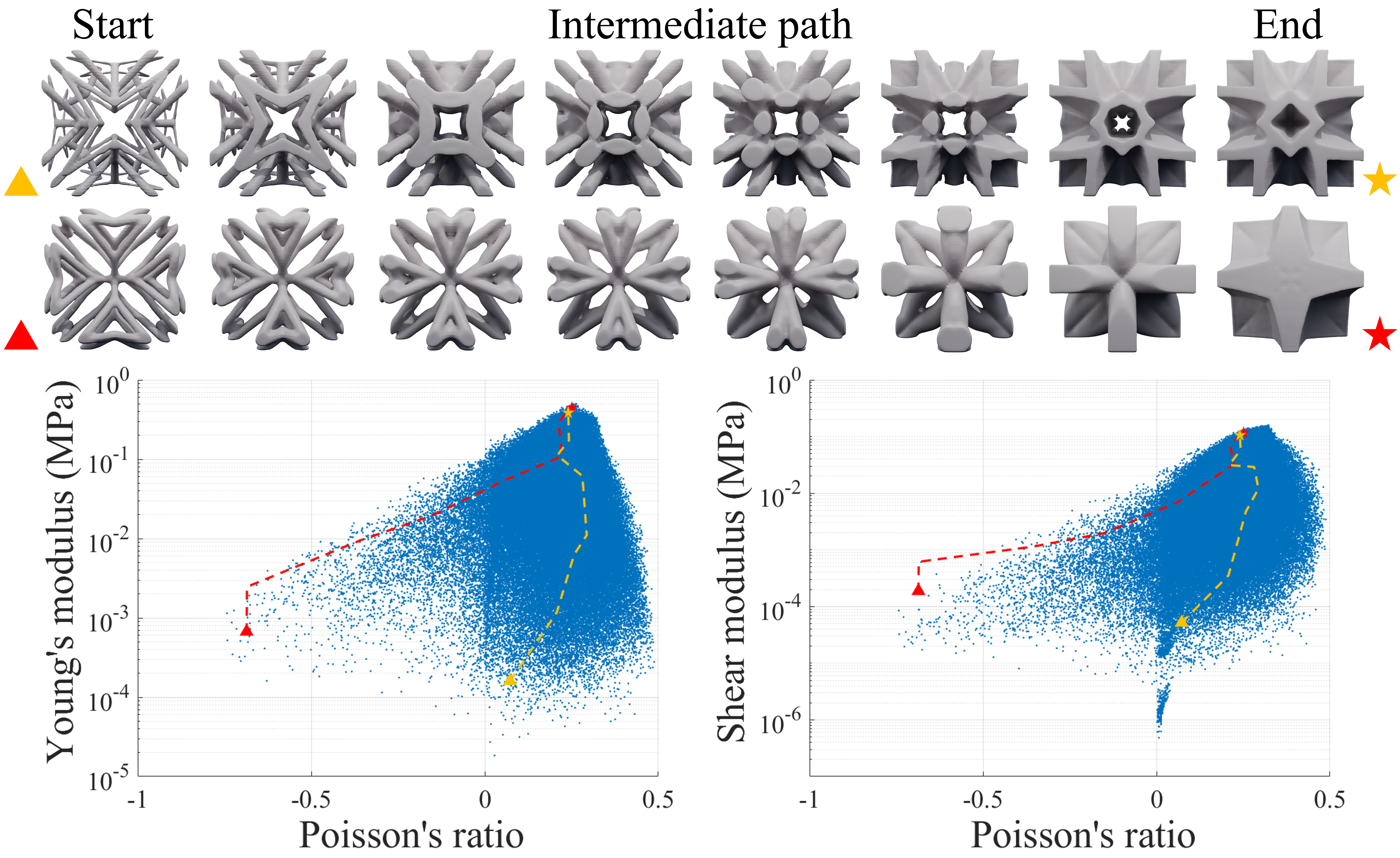}
    \caption{Interpolation using different properties.
    The start models are truss structures with low Young’s moduli and small volume fractions, while the end models are plate structures with high Young’s moduli and larger volume fractions.
    Initially, the interpolated models retain their truss topology, increasing the volume fraction to achieve higher Young’s modulus values.
    Gradually, the structures transition into plate configurations, ultimately forming plate structures with significantly higher Young’s moduli.
    }
    \label{fig:interpolation}
\end{figure}

\begin{figure}[tb]
    \centering
    \includegraphics[width=1.0\linewidth]{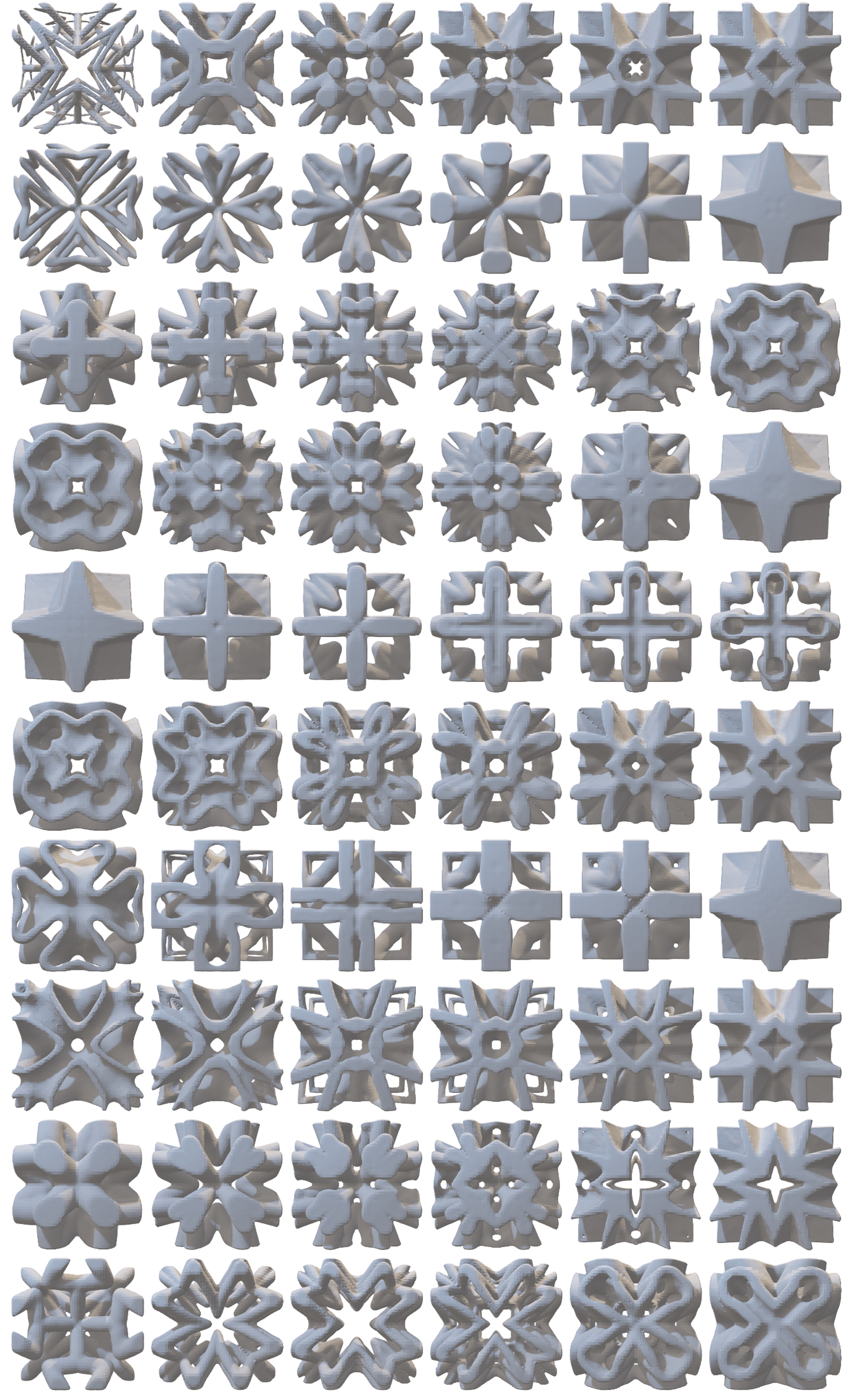}
    \caption{\rev{Interpolation across distinct microstructure families. Each row shows an interpolation sequence between two distinct microstructures: the left-most and right-most shapes in each row are the start and end points. Intermediate shapes are generated by interpolating between them.}}
    \label{fig:supp_interpolation}
\end{figure}

Our approach also enables the generation of novel structures through interpolations, as described in Sec.~\ref{sec:compat}.
Specifically, we performed interpolation experiments on two groups of models with significantly different material properties, as well as several groups of models belonging to distinct microstructure families.
As shown in Fig.~\ref{fig:interpolation} and Fig.~\ref{fig:supp_interpolation}, our method achieves smooth geometric and physical transitions within each group of configurations.

\setlength{\belowcaptionskip}{0pt}

\subsection{Printability}
\begin{figure*}
    \centering
    \includegraphics[width=\linewidth]{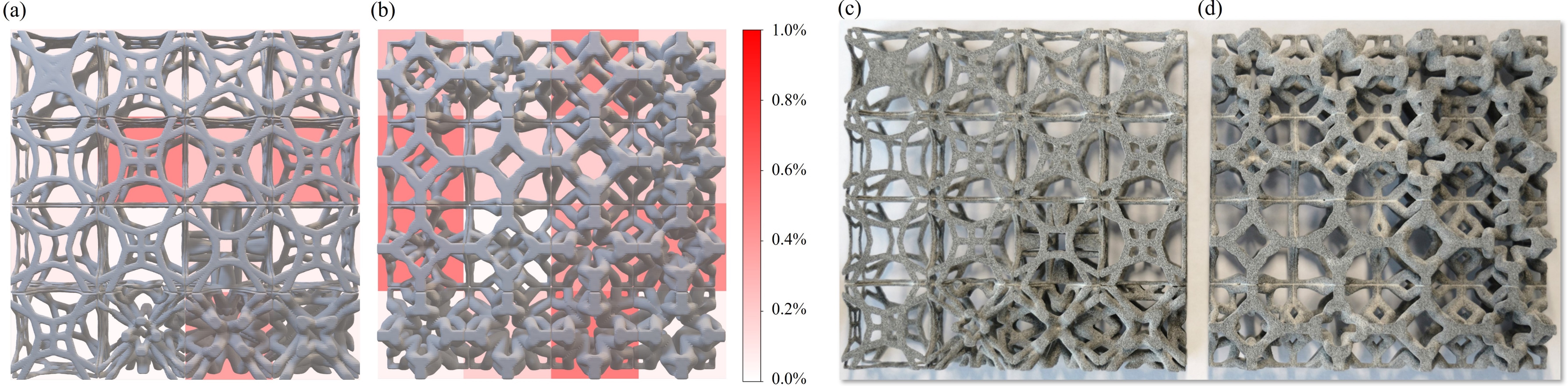}
    \caption{(a) and (b) show two generated microstructures under different printing precisions, along with their property error distributions visualized as heatmaps.
    (c) and (d) present the printed results corresponding to (a) and (b).
    The two test models used a $4 \times 4 \times 1$ grid of microstructures, each with a cell size of 20 mm.
    The experiments were conducted with printer precision settings of \textbf{0.6 mm} and \textbf{1.2 mm}, corresponding to detection resolutions of $32^3$ and $16^3$.
    The smallest feature sizes in the first and second test cases were \textbf{0.62 mm} and \textbf{1.25 mm}, respectively, with property errors of \textbf{0.2\%} and \textbf{0.4\%}.
    For the first structure, the smallest feature size was 0.62 mm, while the second structure exhibited a minimum feature size of 1.25 mm.}
    \label{fig:print}
\end{figure*}

We tested the ability of our method to generate printable objects at different printing precisions.
Finer printing resolution allows for more precise error control.
Experiments show that at both printing resolutions, our method can still generate printable structures that meet the required property specifications (Fig.~\ref{fig:print}).

\subsection{Heterogeneous Design}
\begin{figure*}
    \centering
    \includegraphics[width=\linewidth]{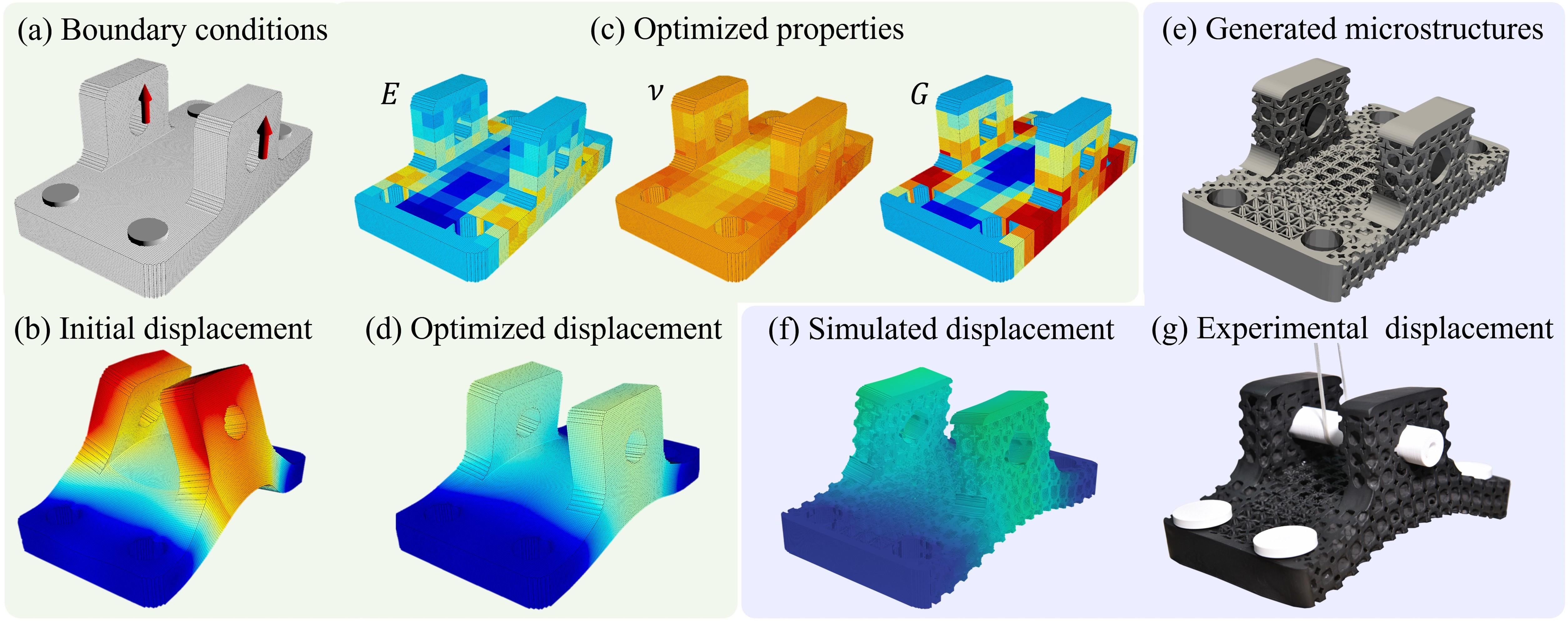}
    \caption{
    \rev{We tested MIND on the pillow bracket model with a 0.01 m grid resolution.
    The base of the model was fixed, and forces were applied to the two handles (a).
    Initial material properties were set to $E = 0.1$, $\nu = 0.25$, and $G = 0.03$, resulting in displacements of \textbf{0.000 m | 0.016 m | 0.039 m} (min, avg, max) (b).
    The design domain was discretized into 392 cells, and we optimized the material properties by constraining the total Young’s modulus while adjusting local stiffness values (c).
    After optimization, the displacement was reduced to \textbf{0.000 m | 0.008 m | 0.032 m} (d).
    The microstructures generated by MIND were then filled in the design region (e).
    The filled model produced a displacement \textbf{of 0.000 m | 0.008 m | 0.034 m} (f), closely matching the optimization results.
    We additionally fabricated the model via 3D printing and conducted physical load testing, which demonstrated performance consistent with the anticipated behavior (f). }
    }
    \label{fig:pillow}
\end{figure*}

To validate MIND in heterogeneous design, we tested it on a pillow bracket model discretized into a 0.01 m grid.
The material properties were optimized to minimize overall displacement.
After optimizing the material distribution, we applied MIND for inverse design and blended the boundaries.
The resulting displacement performance closely aligned with the optimization target, demonstrating MIND’s capability to generate microstructures that meet mechanical requirements while ensuring boundary compatibility (Fig.~\ref{fig:pillow}).

\section{Conclusion and Future Work}
\label{sec:conclusion}

In this paper, we propose a generative model for the inverse design of 3D tileable microstructures.
Our end-to-end model integrates fabrication constraints and generates multiple microstructures with desired physical properties.
Unlike existing methods that rely on specific parameterizations or constrained design spaces, our approach offers a more flexible and general framework.
The proposed latent representation, Holoplane, encodes both geometry and properties in an explicit-implicit hybrid form, improving the alignment between geometry and property distributions.
This results in higher property matching accuracy, enhanced validity control over the generated microstructures, and better boundary compatibility.
The generated microstructures span various architecture types, forming a continuous functional space with diverse morphologies.

The proposed framework has certain limitations that we plan to address in future work.
More fabrication challenges such as self-supportiveness and the absence of closed pockets remain unaddressed and can currently only be ensured through post-filtering.
Incorporating these fabrication constraints into the loss function would be a valuable direction for future investigation, especially given recent progress in this area~\cite{Chen2024,Guo2024PhysComp}.
\rev{
In our study, we tested the pipeline exclusively on structures with cubic symmetry.
Extending it to a broader range of symmetry groups would be interesting (\sreftraining).
Combining it with our interpolation and blending methods may also yield more asymmetric structures. }
We believe our workflow could be expanded to include properties beyond Young's modulus, Poisson's ratio, and shear modulus.
Incorporating properties such as isotropy, thermal conductivity, optical behavior, and others could broaden the range of achievable microstructures and facilitate the design of materials with tailored multi-functionalities.

\begin{acks}
We thank all reviewers for their valuable comments and constructive suggestions.
We are also grateful to Prof. Niloy J. Mitra for his insightful discussions, and to Dr. Lingxin Cao and Yu Xing for their assistance with the simulations and for proofreading the paper.
This work is supported in part by grants from National Key Research and Development Program of China (Grant No. 2024YFB3309500), Key Research Project of Zhejiang Lab (Grant No. 3700-3AA240100), and National Natural Science Foundation of China (Grant Nos. 62472258, U23A20312, 62472257).
\end{acks}

\bibliographystyle{ACM-Reference-Format}
\bibliography{MIND}

@inproceedings{ion2016Metamaterial,
  title = {Metamaterial Mechanisms},
  booktitle = {Proceedings of the 29th {Annual Symposium} on {User Interface Software} and {Technology} - {UIST} '16},
  author = {Ion, Alexandra and Frohnhofen, Johannes and Wall, Ludwig and Kovacs, Robert and Alistar, Mirela and Lindsay, Jack and Lopes, Pedro and Chen, Hsiang-Ting and Baudisch, Patrick},
  year = {2016},
  month = oct,
  pages = {529--539},
  publisher = {ACM Press},
  address = {Tokyo Japan},
  doi = {10.1145/2984511.2984540},
  isbn = {978-1-4503-4189-9},
  langid = {english}
}

@inproceedings{ion2018Metamaterial,
  title = {Metamaterial Textures},
  booktitle = {Proceedings of the 2018 {CHI Conference} on {Human Factors} in {Computing Systems} - {CHI} '18},
  author = {Ion, Alexandra and Kovacs, Robert and Schneider, Oliver S. and Lopes, Pedro and Baudisch, Patrick},
  year = {2018},
  month = apr,
  pages = {1--12},
  publisher = {ACM Press},
  address = {Montreal QC Canada},
  doi = {10.1145/3173574.3173910},
  isbn = {978-1-4503-5620-6},
  langid = {english}
}

@Article{panetta2017Worstcase,
  author    = {Panetta, Julian and Rahimian, Abtin and Zorin, Denis},
  journal   = {{ACM} Transactions on Graphics},
  title     = {Worst-Case Stress Relief for Microstructures},
  year      = {2017},
  issn      = {0730-0301},
  month     = jul,
  number    = {4},
  pages     = {122:1--122:16},
  volume    = {36},
  address   = {New York, NY, USA},
  doi       = {10.1145/3072959.3073649},
  publisher = {Association for Computing Machinery},
}

@Article{Sun2023,
  author    = {Sun, Bingteng and Yan, Xin and Liu, Peiqing and Xia, Yang and Lu, Lin},
  journal   = {Additive Manufacturing},
  title     = {Parametric plate lattices: Modeling and optimization of plate lattices with superior mechanical properties},
  year      = {2023},
  issn      = {2214-8604},
  month     = jun,
  pages     = {103626},
  volume    = {72},
  doi       = {10.1016/j.addma.2023.103626},
  publisher = {Elsevier BV},
}

@Article{Xu2023,
  author    = {Xu, Yonglai and Pan, Hao and Wang, Ruonan and Du, Qiang and Lu, Lin},
  journal   = {Additive Manufacturing},
  title     = {New families of triply periodic minimal surface-like shell lattices},
  year      = {2023},
  issn      = {2214-8604},
  month     = sep,
  pages     = {103779},
  volume    = {77},
  doi       = {10.1016/j.addma.2023.103779},
  publisher = {Elsevier BV},
}

@article{yu2014flat,
  title={Flat optics with designer metasurfaces},
  author={Yu, Nanfang and Capasso, Federico},
  journal={Nature materials},
  volume={13},
  number={2},
  pages={139--150},
  year={2014},
  publisher={Nature Publishing Group UK London}
}

@article{andreassen2014Design,
  title = {Design of Manufacturable {3D} Extremal Elastic Microstructure},
  author = {Andreassen, Erik and Lazarov, Boyan S. and Sigmund, Ole},
  year = {2014},
  month = feb,
  journal = {Mechanics of Materials},
  volume = {69},
  number = {1},
  pages = {1--10},
  issn = {01676636},
  doi = {10.1016/j.mechmat.2013.09.018},  
}

@article{bastek2022Inverting,
  title = {Inverting the Structure\textendash Property Map of Truss Metamaterials by Deep Learning},
  author = {Bastek, Jan-Hendrik and Kumar, Siddhant and Telgen, Bastian and Glaesener, Rapha{\"e}l N. and Kochmann, Dennis M.},
  year = {2022},
  month = jan,
  journal = {Proceedings of the National Academy of Sciences},
  volume = {119},
  number = {1},
  pages = {e2111505119},
  issn = {0027-8424, 1091-6490},
  doi = {10.1073/pnas.2111505119},
  }

@article{bonatti2019Mechanical,
  title = {Mechanical Performance of Additively-Manufactured Anisotropic and Isotropic Smooth Shell-Lattice Materials: {Simulations} \& Experiments},
  shorttitle = {Mechanical Performance of Additively-Manufactured Anisotropic and Isotropic Smooth Shell-Lattice Materials},
  author = {Bonatti, Colin and Mohr, Dirk},
  year = {2019},
  month = jan,
  journal = {Journal of the Mechanics and Physics of Solids},
  volume = {122},
  pages = {1--26},
  issn = {00225096},
  doi = {10.1016/j.jmps.2018.08.022},
  langid = {english}
}

@article{bostanabad2019Globally,
  title = {Globally {Approximate Gaussian Processes} for {Big Data With Application} to {Data-Driven Metamaterials Design}},
  author = {Bostanabad, Ramin and Chan, Yu-Chin and Wang, Liwei and Zhu, Ping and Chen, Wei},
  year = {2019},
  month = nov,
  journal = {Journal of Mechanical Design},
  volume = {141},
  number = {11},
  pages = {111402},
  issn = {1050-0472, 1528-9001},
  doi = {10.1115/1.4044257},  
}

@article{choi2016Nonlinear,
  title = {Nonlinear Analysis of the Poisson's Ratio of Negative Poisson's Ratio Foams},
  author = {Choi, J. B. and Lakes, R. S.},
  year = {2016},
  month = jul,
  journal = {Journal of Composite Materials},
  volume = {29},
  number = {1},
  pages = {113--128},
  publisher = {Sage PublicationsSage CA: Thousand Oaks, CA},
  issn = {0021-9983},
  doi = {10.1177/002199839502900106},  
}

@article{dong2018149,
  title = {A 149 Line Homogenization Code for Three-Dimensional Cellular Materials Written in Matlab},
  author = {Dong, Guoying and Tang, Yunlong and Zhao, Yaoyao Fiona},
  year = {2018},
  month = jul,
  journal = {Journal of Engineering Materials and Technology},
  volume = {141},
  number = {1},
  publisher = {American Society of Mechanical Engineers Digital Collection},
  doi = {10.1115/1.4040555}
}

@article{goodfellow2020Generative,
  title = {Generative Adversarial Networks},
  author = {Goodfellow, Ian and {Pouget-Abadie}, Jean and Mirza, Mehdi and Xu, Bing and {Warde-Farley}, David and Ozair, Sherjil and Courville, Aaron and Bengio, Yoshua},
  year = {2020},
  month = oct,
  journal = {Communications of the ACM},
  volume = {63},
  number = {11},
  pages = {139--144},
  issn = {0001-0782, 1557-7317},
  doi = {10.1145/3422622},
  
}

@Article{Hsu2020,
  author    = {Tim Hsu and William K. Epting and Hokon Kim and Harry W. Abernathy and Gregory A. Hackett and Anthony D. Rollett and Paul A. Salvador and Elizabeth A. Holm},
  journal   = {{JOM}},
  title     = {Microstructure Generation via Generative Adversarial Network for Heterogeneous, Topologically Complex 3D Materials},
  year      = {2020},
  month     = {dec},
  number    = {1},
  pages     = {90--102},
  volume    = {73},
  doi       = {10.1007/s11837-020-04484-y},
  publisher = {Springer Science and Business Media {LLC}},
}

@article{hu2019Lightweight,
  title = {A Lightweight Methodology of {3D} Printed Objects Utilizing Multi-Scale Porous Structures},
  author = {Hu, Jiangbei and Wang, Shengfa and Wang, Yi and Li, Fengqi and Luo, Zhongxuan},
  year = {2019},
  month = jun,
  journal = {The Visual Computer},
  volume = {35},
  number = {6},
  pages = {949--959},
  issn = {1432-2315},
  doi = {10.1007/s00371-019-01672-z},  
}

@article{hu2020Cellular,
  title = {Cellular Structure Design Based on Free Material Optimization under Connectivity Control},
  author = {Hu, Jingqiao and Li, Ming and Yang, Xingtong and Gao, Shuming},
  year = {2020},
  month = oct,
  journal = {Computer-Aided Design},
  volume = {127},
  pages = {102854},
  issn = {0010-4485},
  doi = {10.1016/j.cad.2020.102854},
  
}

@inproceedings{li2018Deep,
  title = {A {Deep Adversarial Learning Methodology} for {Designing Microstructural Material Systems}},
  booktitle = {Volume {2B}: 44th {Design Automation Conference}},
  author = {Li, Xiaolin and Yang, Zijiang and Brinson, L. Catherine and Choudhary, Alok and Agrawal, Ankit and Chen, Wei},
  year = {2018},
  month = aug,
  pages = {V02BT03A008},
  publisher = {American Society of Mechanical Engineers},
  address = {Quebec City, Quebec, Canada},
  doi = {10.1115/DETC2018-85633},
  isbn = {978-0-7918-5176-0},
}

@article{ling2019Mechanical,
  title = {Mechanical Behaviour of Additively-Manufactured Polymeric Octet-Truss Lattice Structures under Quasi-Static and Dynamic Compressive Loading},
  author = {Ling, Chen and Cernicchi, Alessandro and Gilchrist, Michael D. and Cardiff, Philip},
  year = {2019},
  month = jan,
  journal = {Materials \& Design},
  volume = {162},
  pages = {106--118},
  publisher = {Elsevier},
  issn = {0264-1275},
  doi = {10.1016/j.matdes.2018.11.035},
}

@article{liu2021MPP,
  title = {Mechanical Property Profiles of Microstructures via Asymptotic Homogenization},
  author = {Liu, Peiqing and Liu, An and Peng, Hao and Tian, Lihao and Liu, Jikai and Lu, Lin},
  year = {2021},
  journal = {Computers \& Graphics},
  volume = {100},
  pages = {106--115},
  publisher = {Elsevier}
}

@article{liu2022Parametric,
  title = {Parametric Shell Lattice with Tailored Mechanical Properties},
  author = {Liu, Peiqing and Sun, Bingteng and Liu, Jikai and Lu, Lin},
  year = {2022},
  month = dec,
  journal = {Additive Manufacturing},
  volume = {60},
  pages = {103258},
  issn = {22148604},
  doi = {10.1016/j.addma.2022.103258},
  langid = {english}
}

@article{lu2014Buildtolast,
  title = {Build-to-Last: Strength to Weight {3D} Printed Objects},
  author = {Lu, Lin and Sharf, Andrei and Zhao, Haisen and Wei, Yuan and Fan, Qingnan and Chen, Xuelin and Savoye, Yann and Tu, Changhe and {Cohen-Or}, Daniel and Chen, Baoquan},
  year = {2014},
  month = jul,
  journal = {{ACM} Transactions on Graphics},
  volume = {33},
  number = {4},
  pages = {97:1--97:10},
  publisher = {ACM New York, NY, USA},
  issn = {0730-0301},
  doi = {10.1145/2601097.2601168}
}

@article{martinez2016Procedural,
  title = {Procedural Voronoi Foams for Additive Manufacturing},
  author = {Mart{\'i}nez, Jon{\`a}s and Dumas, J{\'e}r{\'e}mie and Lefebvre, Sylvain},
  year = {2016},
  month = jul,
  journal = {{ACM} Transactions on Graphics},
  volume = {35},
  number = {4},
  pages = {44:1--44:12},
  publisher = {ACM},
  address = {New York, NY, USA},
  issn = {0730-0301},
  doi = {10.1145/2897824.2925922},
  acmid = {2925922},
  articleno = {44}
}

@article{martinez2017Orthotropic,
  title = {Orthotropic {K-nearest} Foams for Additive Manufacturing},
  author = {Mart{\'i}nez, Jon{\`a}s and Song, Haichuan and Dumas, J{\'e}r{\'e}mie and Lefebvre, Sylvain},
  year = {2017},
  month = jul,
  journal = {{ACM} Transactions on Graphics},
  volume = {36},
  number = {4},
  pages = {121:1--121:12},
  publisher = {ACM},
  address = {New York, NY, USA},
  issn = {0730-0301},
  doi = {10.1145/3072959.3073638},
}

@article{martinez2018Polyhedral,
  title = {Polyhedral Voronoi Diagrams for Additive Manufacturing},
  author = {Mart{\'i}nez, Jon{\`a}s and Hornus, Samuel and Song, Haichuan and Lefebvre, Sylvain},
  year = {2018},
  month = jul,
  journal = {{ACM} Transactions on Graphics},
  volume = {37},
  number = {4},
  pages = {129:1--129:15},
  publisher = {ACM},
  address = {New York, NY, USA},
  issn = {0730-0301},
  doi = {10.1145/3197517.3201343}, 
}

@misc{KingmaVAE,
  author        = {Diederik P Kingma and Max Welling},
  title         = {Auto-Encoding Variational Bayes},
  year          = {2013},
  month         = dec,
  archiveprefix = {arXiv},
  eprint        = {1312.6114},
  primaryclass  = {stat.ML},
}

@article{nazir2019Stateoftheart,
  title = {A State-of-the-Art Review on Types, Design, Optimization, and Additive Manufacturing of Cellular Structures},
  author = {Nazir, Aamer and Abate, Kalayu Mekonen and Kumar, Ajeet and Jeng, Jeng-Ywan},
  year = {2019},
  month = oct,
  journal = {The International Journal of Advanced Manufacturing Technology},
  volume = {104},
  number = {9},
  pages = {3489--3510},
  publisher = {Springer London},
  issn = {1433-3015},
  doi = {10.1007/s00170-019-04085-3},
  copyright = {2019 Springer-Verlag London Ltd., part of Springer Nature},
  }

@article{noguchi2021Stochastic,
  title = {Stochastic Characterization and Reconstruction of Material Microstructures for Establishment of Process-Structure-Property Linkage Using the Deep Generative Model},
  author = {Noguchi, Satoshi and Inoue, Junya},
  year = {2021},
  month = aug,
  journal = {Physical Review E},
  volume = {104},
  number = {2},
  pages = {025302},
  issn = {2470-0045, 2470-0053},
  doi = {10.1103/PhysRevE.104.025302},  
}

@article{overvelde2016Threedimensional,
  title = {A Three-Dimensional Actuated Origami-Inspired Transformable Metamaterial with Multiple Degrees of Freedom},
  author = {Overvelde, Johannes T. B. and {de Jong}, Twan A. and Shevchenko, Yanina and Becerra, Sergio A. and Whitesides, George M. and Weaver, James C. and Hoberman, Chuck and Bertoldi, Katia},
  year = {2016},
  month = mar,
  journal = {Nature Communications},
  volume = {7},
  number = {1},
  pages = {1--8},
  publisher = {Nature Publishing Group},
  issn = {2041-1723},
  doi = {10.1038/ncomms10929},
  copyright = {2016 The Author(s)},
  }

@article{panetta2015Elastic,
  title = {Elastic Textures for Additive Fabrication},
  author = {Panetta, Julian and Zhou, Qingnan and Malomo, Luigi and Pietroni, Nico and Cignoni, Paolo and Zorin, Denis},
  year = {2015},
  month = jul,
  journal = {{ACM} Transactions on Graphics},
  volume = {34},
  number = {4},
  pages = {135:1--135:12},
  publisher = {Association for Computing Machinery (ACM)},
  issn = {0730-0301},
  doi = {10.1145/2766937}
}

@article{tang2023beyond,
  title={Beyond Chainmail: Computational Modeling of Discrete Interlocking Materials},
  author={Tang, Pengbin and Coros, Stelian and Thomaszewski, Bernhard},
  journal={ACM Transactions on Graphics (TOG)},
  volume={42},
  number={4},
  pages={1--12},
  year={2023},
  publisher={ACM New York, NY, USA}
}

@article{schumacher2015Microstructures,
  title = {Microstructures to Control Elasticity in {3D} Printing},
  author = {Schumacher, Christian and Bickel, Bernd and Rys, Jan and Marschner, Steve and Daraio, Chiara and Gross, Markus},
  year = {2015},
  month = jul,
  journal = {{ACM} Transactions on Graphics},
  volume = {34},
  number = {4},
  pages = {136:1--136:13},
  publisher = {Association for Computing Machinery (ACM)},
  issn = {0730-0301},
  doi = {10.1145/2766926}
}

@inproceedings{tian2020Organic,
  title = {Organic Open-Cell Porous Structure Modeling},
  booktitle = {Symposium on {Computational Fabrication}},
  author = {Tian, Lihao and Lu, Lin and Chen, Weikai and Xia, Yang and Wang, Charlie C. L. and Wang, Wenping},
  year = {2020},
  month = nov,
  pages = {1--12},
  publisher = {ACM},
  address = {Virtual Event USA},
  doi = {10.1145/3424630.3425414},
  isbn = {978-1-4503-8170-3},
  langid = {english}
}

@incollection{torquato2005Microstructure,
  title = {Microstructure Optimization},
  booktitle = {Handbook of {Materials Modeling}},
  author = {Torquato, S.},
  editor = {Yip, Sidney},
  year = {2005},
  pages = {2379--2396},
  publisher = {Springer Netherlands},
  address = {Dordrecht},
  doi = {10.1007/978-1-4020-3286-8_124},
  isbn = {978-1-4020-3287-5 978-1-4020-3286-8},
  langid = {english}
}

@article{tozoni2020Lowparametric,
  title = {A Low-Parametric Rhombic Microstructure Family for Irregular Lattices},
  author = {Tozoni, Davi Colli and Dumas, J{\'e}r{\'e}mie and Jiang, Zhongshi and Panetta, Julian and Panozzo, Daniele and Zorin, Denis},
  year = {2020},
  month = aug,
  journal = {{ACM} Transactions on Graphics},
  volume = {39},
  number = {4},
  pages = {101--1},
  publisher = {Association for Computing Machinery (ACM)},
  issn = {0730-0301, 1557-7368},
  doi = {10.1145/3386569.3392451},
  langid = {english}
}

@article{wang2020Datadriven,
  title = {Data-Driven Metamaterial Design with {Laplace-Beltrami} Spectrum as ``Shape-{{DNA}}''},
  author = {Wang, Liwei and Chan, Yu-Chin and Liu, Zhao and Zhu, Ping and Chen, Wei},
  year = {2020},
  month = jun,
  journal = {Structural and Multidisciplinary Optimization},
  volume = {61},
  number = {6},
  pages = {2613--2628},
  issn = {1615-147X, 1615-1488},
  doi = {10.1007/s00158-020-02523-5},
}

@article{wang2020Deep,
  title = {Deep Generative Modeling for Mechanistic-Based Learning and Design of Metamaterial Systems},
  author = {Wang, Liwei and Chan, Yu-Chin and Ahmed, Faez and Liu, Zhao and Zhu, Ping and Chen, Wei},
  year = {2020},
  month = dec,
  journal = {Computer Methods in Applied Mechanics and Engineering},
  volume = {372},
  pages = {113377},
  issn = {00457825},
  doi = {10.1016/j.cma.2020.113377},
}

@article{wang2020Quasiperiodica,
  title = {Quasiperiodic Mechanical Metamaterials with Extreme Isotropic Stiffness},
  author = {Wang, Yiqiang and Sigmund, Ole},
  year = {2020},
  month = jan,
  journal = {Extreme Mechanics Letters},
  volume = {34},
  pages = {100596},
  issn = {23524316},
  doi = {10.1016/j.eml.2019.100596},
}

@article{wang2021DataDriven,
  title = {Data-{Driven Topology Optimization With Multiclass Microstructures Using Latent Variable Gaussian Process}},
  author = {Wang, Liwei and Tao, Siyu and Zhu, Ping and Chen, Wei},
  year = {2021},
  month = mar,
  journal = {Journal of Mechanical Design},
  volume = {143},
  number = {3},
  pages = {031708},
  issn = {1050-0472, 1528-9001},
  doi = {10.1115/1.4048628},
}

@article{wang2022IHGAN,
  title = {{IH-GAN}: {A} Conditional Generative Model for Implicit Surface-Based Inverse Design of Cellular Structures},
  shorttitle = {{IH-GAN}},
  author = {Wang, Jun and Chen, Wei (Wayne) and Da, Daicong and Fuge, Mark and Rai, Rahul},
  year = {2022},
  month = jun,
  journal = {Computer Methods in Applied Mechanics and Engineering},
  volume = {396},
  pages = {115060},
  issn = {00457825},
  doi = {10.1016/j.cma.2022.115060},  
}

@Article{wu2018Infill,
  author    = {Jun Wu and Niels Aage and Rudiger Westermann and Ole Sigmund},
  journal   = {{IEEE} Transactions on Visualization and Computer Graphics},
  title     = {Infill Optimization for Additive Manufacturing-Approaching Bone-Like Porous Structures},
  year      = {2018},
  month     = {feb},
  number    = {2},
  pages     = {1127--1140},
  volume    = {24},
  doi       = {10.1109/tvcg.2017.2655523},
  publisher = {Institute of Electrical and Electronics Engineers ({IEEE})},
}

@article{yan2020Strong,
  title = {Strong {3D} Printing by {TPMS} Injection},
  author = {Yan, Xin and Rao, Cong and Lu, Lin and Sharf, Andrei and Zhao, Haisen and Chen, Baoquan},
  year = {2020},
  month = oct,
  journal = {IEEE Transactions on Visualization and Computer Graphics},
  volume = {26},
  number = {10},
  pages = {3037--3050},
  issn = {1941-0506},
  doi = {10.1109/tvcg.2019.2914044},
  }

@article{zheng2021Controllable,
  title = {Controllable Inverse Design of Auxetic Metamaterials Using Deep Learning},
  author = {Zheng, Xiaoyang and Chen, Ta-Te and Guo, Xiaofeng and Samitsu, Sadaki and Watanabe, Ikumu},
  year = {2021},
  month = dec,
  journal = {Materials \& Design},
  volume = {211},
  pages = {110178},
  issn = {02641275},
  doi = {10.1016/j.matdes.2021.110178},  
}

@article{Ding2021,
title = {STL-free design and manufacturing paradigm for high-precision powder bed fusion},
journal = {CIRP Annals},
volume = {70},
number = {1},
pages = {167-170},
year = {2021},
issn = {0007-8506},
doi = {https://doi.org/10.1016/j.cirp.2021.03.012},
url = {https://www.sciencedirect.com/science/article/pii/S0007850621000123},
author = {Junhao Ding and Qiang Zou and Shuo Qu and Paulo Bartolo and Xu Song and Charlie C.L. Wang}
}

@article{Gao2018,
title = {On topology optimization and canonical duality method},
journal = {Computer Methods in Applied Mechanics and Engineering},
volume = {341},
pages = {249-277},
year = {2018},
issn = {0045-7825},
doi = {https://doi.org/10.1016/j.cma.2018.06.027},
url = {https://www.sciencedirect.com/science/article/pii/S0045782518303256},
author = {David Yang Gao}
}

@Article{Sigmund1994,
  author    = {Ole Sigmund},
  journal   = {International Journal of Solids and Structures},
  title     = {Materials with prescribed constitutive parameters: An inverse homogenization problem},
  year      = {1994},
  month     = {sep},
  number    = {17},
  pages     = {2313--2329},
  volume    = {31},
  doi       = {10.1016/0020-7683(94)90154-6},
  publisher = {Elsevier {BV}},
}

@Article{zhang2023optimized,
  author    = {Zhang, Di and Zhai, Xiaoya and Liu, Ligang and Fu, Xiao-Ming},
  journal   = {Structural and Multidisciplinary Optimization},
  title     = {An optimized, easy-to-use, open-source GPU solver for large-scale inverse homogenization problems},
  year      = {2023},
  issn      = {1615-1488},
  month     = sep,
  number    = {9},
  volume    = {66},
  doi       = {10.1007/s00158-023-03657-y},
  publisher = {Springer Science and Business Media LLC},
}

@Article{Zhang2024Clay,
  author    = {Zhang, Longwen and Wang, Ziyu and Zhang, Qixuan and Qiu, Qiwei and Pang, Anqi and Jiang, Haoran and Yang, Wei and Xu, Lan and Yu, Jingyi},
  journal   = {ACM Transactions on Graphics},
  title     = {CLAY: A Controllable Large-scale Generative Model for Creating High-quality 3D Assets},
  year      = {2024},
  issn      = {1557-7368},
  month     = jul,
  number    = {4},
  pages     = {1--20},
  volume    = {43},
  doi       = {10.1145/3658146},
  publisher = {Association for Computing Machinery (ACM)},
}

@Article{guo2024physcomp,
  author  = {Guo, Minghao and Wang, Bohan and Ma, Pingchuan and Zhang, Tianyuan and Owens, Crystal Elaine and Gan, Chuang and Tenenbaum, Joshua B. and He, Kaiming and Matusik, Wojciech},
  journal = {arXiv preprint arXiv:2405.20510},
  title   = {Physically Compatible 3D Object Modeling from a Single Image},
  year    = {2024},
}

@Article{Chen2024,
  author        = {Chen, Yunuo and Xie, Tianyi and Zong, Zeshun and Li, Xuan and Gao, Feng and Yang, Yin and Wu, Ying Nian and Jiang, Chenfanfu},
  title         = {Atlas3D: Physically Constrained Self-Supporting Text-to-3D for Simulation and Fabrication},
  year          = {2024},
  month         = may,
  archiveprefix = {arXiv},
  copyright     = {Creative Commons Attribution 4.0 International},
  doi           = {10.48550/ARXIV.2405.18515},
  eprint        = {2405.18515},
  primaryclass  = {cs.LG},
  publisher     = {arXiv},
}

@Misc{Yang2024,
  author        = {Yang, Yanyan and Wang, Lili and Zhai, Xiaoya and Chen, Kai and Wu, Wenming and Zhao, Yunkai and Liu, Ligang and Fu, Xiao-Ming},
  month         = jan,
  title         = {Guided Diffusion for Fast Inverse Design of Density-based Mechanical Metamaterials},
  year          = {2024},
  archiveprefix = {arXiv},
  copyright     = {arXiv.org perpetual, non-exclusive license},
  doi           = {10.48550/ARXIV.2401.13570},
  eprint        = {2401.13570},
  primaryclass  = {cs.CE},
  publisher     = {arXiv},
}

@Article{Zheng2023,
  author    = {Zheng, Li and Karapiperis, Konstantinos and Kumar, Siddhant and Kochmann, Dennis M.},
  journal   = {Nature Communications},
  title     = {Unifying the design space and optimizing linear and nonlinear truss metamaterials by generative modeling},
  year      = {2023},
  issn      = {2041-1723},
  month     = nov,
  number    = {1},
  volume    = {14},
  doi       = {10.1038/s41467-023-42068-x},
  publisher = {Springer Science and Business Media LLC},
}

@Article{Huang2024,
  author    = {Huang, Zizhou and Panozzo, Daniele and Zorin, Denis},
  journal   = {ACM Transactions on Graphics},
  title     = {Optimized shock-protecting microstructures},
  year      = {2024},
  issn      = {1557-7368},
  month     = nov,
  number    = {6},
  pages     = {1--21},
  volume    = {43},
  doi       = {10.1145/3687765},
  publisher = {Association for Computing Machinery (ACM)},
}

@Article{Ha2023,
  author    = {Ha, Chan Soo and Yao, Desheng and Xu, Zhenpeng and Liu, Chenang and Liu, Han and Elkins, Daniel and Kile, Matthew and Deshpande, Vikram and Kong, Zhenyu and Bauchy, Mathieu and Zheng, Xiaoyu},
  journal   = {Nature Communications},
  title     = {Rapid inverse design of metamaterials based on prescribed mechanical behavior through machine learning},
  year      = {2023},
  issn      = {2041-1723},
  month     = sep,
  number    = {1},
  volume    = {14},
  doi       = {10.1038/s41467-023-40854-1},
  publisher = {Springer Science and Business Media LLC},
}

@Article{Li2023,
  author    = {Li, Yue and Coros, Stelian and Thomaszewski, Bernhard},
  journal   = {ACM Transactions on Graphics},
  title     = {Neural Metamaterial Networks for Nonlinear Material Design},
  year      = {2023},
  issn      = {1557-7368},
  month     = dec,
  number    = {6},
  pages     = {1--13},
  volume    = {42},
  doi       = {10.1145/3618325},
  publisher = {Association for Computing Machinery (ACM)},
}

@Article{Coelho2007,
  author    = {Coelho, P. G. and Fernandes, P. R. and Guedes, J. M. and Rodrigues, H. C.},
  journal   = {Structural and Multidisciplinary Optimization},
  title     = {A hierarchical model for concurrent material and topology optimisation of three-dimensional structures},
  year      = {2007},
  issn      = {1615-1488},
  month     = jun,
  number    = {2},
  pages     = {107--115},
  volume    = {35},
  doi       = {10.1007/s00158-007-0141-3},
  publisher = {Springer Science and Business Media LLC},
}

@ARTICLE{Deng2024neural,
  author={Deng, Zhi and Xiao, Haoyao and Lang, Yining and Feng, Hao and Zhang, Juyong},
  journal={Computational Visual Media}, 
  title={Multi-scale hash encoding based neural geometry representation}, 
  year={2024},
  volume={10},
  number={3},
  pages={453-470},
  doi={10.1007/s41095-023-0340-x}}

@Article{Lakes1987,
  author    = {Lakes, Roderic},
  journal   = {Science},
  title     = {Foam Structures with a Negative Poisson’s Ratio},
  year      = {1987},
  issn      = {1095-9203},
  month     = feb,
  number    = {4792},
  pages     = {1038--1040},
  volume    = {235},
  doi       = {10.1126/science.235.4792.1038},
  publisher = {American Association for the Advancement of Science (AAAS)},
}

@Article{Wang2024,
  author    = {Wang, Haoyu and Du, Zongliang and Feng, Fuyong and Kang, Zhong and Tang, Shan and Guo, Xu},
  journal   = {Computer Methods in Applied Mechanics and Engineering},
  title     = {DiffMat: Data-driven inverse design of energy-absorbing metamaterials using diffusion model},
  year      = {2024},
  issn      = {0045-7825},
  month     = dec,
  pages     = {117440},
  volume    = {432},
  doi       = {10.1016/j.cma.2024.117440},
  publisher = {Elsevier BV},
}

@Article{Lee2023,
  author    = {Lee, Doksoo and Chen, Wei (Wayne) and Wang, Liwei and Chan, Yu‐Chin and Chen, Wei},
  journal   = {Advanced Materials},
  title     = {Data‐Driven Design for Metamaterials and Multiscale Systems: A Review},
  year      = {2023},
  issn      = {1521-4095},
  month     = dec,
  number    = {8},
  volume    = {36},
  doi       = {10.1002/adma.202305254},
  publisher = {Wiley},
}

@Article{Bastek2023,
  author    = {Bastek, Jan-Hendrik and Kochmann, Dennis M.},
  journal   = {Nature Machine Intelligence},
  title     = {Inverse design of nonlinear mechanical metamaterials via video denoising diffusion models},
  year      = {2023},
  issn      = {2522-5839},
  month     = dec,
  number    = {12},
  pages     = {1466--1475},
  volume    = {5},
  doi       = {10.1038/s42256-023-00762-x},
  publisher = {Springer Science and Business Media LLC},
}

@Article{Vlassis2023,
  author    = {Vlassis, Nikolaos N. and Sun, WaiChing},
  journal   = {Computer Methods in Applied Mechanics and Engineering},
  title     = {Denoising diffusion algorithm for inverse design of microstructures with fine-tuned nonlinear material properties},
  year      = {2023},
  issn      = {0045-7825},
  month     = aug,
  pages     = {116126},
  volume    = {413},
  doi       = {10.1016/j.cma.2023.116126},
  publisher = {Elsevier BV},
}

@Article{Liu2022,
  author    = {Liu, Ke and Sun, Rachel and Daraio, Chiara},
  journal   = {Science},
  title     = {Growth rules for irregular architected materials with programmable properties},
  year      = {2022},
  issn      = {1095-9203},
  month     = aug,
  number    = {6609},
  pages     = {975--981},
  volume    = {377},
  doi       = {10.1126/science.abn1459},
  publisher = {American Association for the Advancement of Science (AAAS)},
}

@Article{Kumar2020,
  author    = {Kumar, Siddhant and Tan, Stephanie and Zheng, Li and Kochmann, Dennis M.},
  journal   = {npj Computational Materials},
  title     = {Inverse-designed spinodoid metamaterials},
  year      = {2020},
  issn      = {2057-3960},
  month     = jun,
  number    = {1},
  volume    = {6},
  doi       = {10.1038/s41524-020-0341-6},
  publisher = {Springer Science and Business Media LLC},
}

@Article{Li2019,
  author    = {Li, Dawei and Dai, Ning and Tang, Yunlong and Dong, Guoying and Zhao, Yaoyao Fiona},
  journal   = {Journal of Mechanical Design},
  title     = {Design and Optimization of Graded Cellular Structures With Triply Periodic Level Surface-Based Topological Shapes},
  year      = {2019},
  issn      = {1528-9001},
  month     = mar,
  number    = {7},
  volume    = {141},
  doi       = {10.1115/1.4042617},
  publisher = {ASME International},
}

@Article{Li2020,
  author    = {Li, Xiang and Ning, Shaowu and Liu, Zhanli and Yan, Ziming and Luo, Chengcheng and Zhuang, Zhuo},
  journal   = {Computer Methods in Applied Mechanics and Engineering},
  title     = {Designing Phononic Crystal with Anticipated Band Gap through a Deep Learning Based Data-Driven Method},
  year      = {2020},
  issn      = {00457825},
  month     = apr,
  pages     = {112737},
  volume    = {361},
  doi       = {10.1016/j.cma.2019.112737},
  publisher = {Elsevier BV},
}

@Article{Liu2018,
  author    = {Liu, Dianjing and Tan, Yixuan and Khoram, Erfan and Yu, Zongfu},
  journal   = {ACS Photonics},
  title     = {Training Deep Neural Networks for the Inverse Design of Nanophotonic Structures},
  year      = {2018},
  issn      = {2330-4022},
  month     = feb,
  number    = {4},
  pages     = {1365--1369},
  volume    = {5},
  doi       = {10.1021/acsphotonics.7b01377},
  publisher = {American Chemical Society (ACS)},
}

@InProceedings{Ho2020DDPM,
  author    = {Ho, Jonathan and Jain, Ajay and Abbeel, Pieter},
  booktitle = {Proceedings of the 34th International Conference on Neural Information Processing Systems},
  title     = {Denoising Diffusion Probabilistic Models},
  year      = {2020},
  address   = {Red Hook, NY, USA},
  publisher = {Curran Associates Inc.},
  series    = {NIPS'20},
  articleno = {574},
  isbn      = {9781713829546},
  location  = {Vancouver, BC, Canada},
  numpages  = {12},
}

@Article{Askari2020,
  author    = {Askari, Meisam and Hutchins, David A. and Thomas, Peter J. and Astolfi, Lorenzo and Watson, Richard L. and Abdi, Meisam and Ricci, Marco and Laureti, Stefano and Nie, Luzhen and Freear, Steven and Wildman, Ricky and Tuck, Christopher and Clarke, Matt and Woods, Emma and Clare, Adam T.},
  journal   = {Additive Manufacturing},
  title     = {Additive manufacturing of metamaterials: A review},
  year      = {2020},
  issn      = {2214-8604},
  month     = dec,
  pages     = {101562},
  volume    = {36},
  doi       = {10.1016/j.addma.2020.101562},
  publisher = {Elsevier BV},
}

@InProceedings{Shue2023,
  author    = {Shue, J. Ryan and Chan, Eric Ryan and Po, Ryan and Ankner, Zachary and Wu, Jiajun and Wetzstein, Gordon},
  booktitle = {2023 IEEE/CVF Conference on Computer Vision and Pattern Recognition (CVPR)},
  title     = {3D Neural Field Generation Using Triplane Diffusion},
  year      = {2023},
  month     = jun,
  pages     = {20875--20886},
  publisher = {IEEE},
  doi       = {10.1109/cvpr52729.2023.02000},
}

@InProceedings{chan2022triplane,
  author    = {Chan, Eric R. and Lin, Connor Z. and Chan, Matthew A. and Nagano, Koki and Pan, Boxiao and de Mello, Shalini and Gallo, Orazio and Guibas, Leonidas and Tremblay, Jonathan and Khamis, Sameh and Karras, Tero and Wetzstein, Gordon},
  booktitle = {2022 IEEE/CVF Conference on Computer Vision and Pattern Recognition (CVPR)},
  title     = {Efficient Geometry-aware 3D Generative Adversarial Networks},
  year      = {2022},
  month     = jun,
  pages     = {16102--16112},
  publisher = {IEEE},
  doi       = {10.1109/cvpr52688.2022.01565},
}

@Article{ho2022classifierfree,
  author        = {Ho, Jonathan and Salimans, Tim},
  title         = {Classifier-Free Diffusion Guidance},
  year          = {2022},
  month         = jul,
  archiveprefix = {arXiv},
  copyright     = {arXiv.org perpetual, non-exclusive license},
  doi           = {10.48550/ARXIV.2207.12598},
  eprint        = {2207.12598},
  primaryclass  = {cs.LG},
  publisher     = {arXiv},
}

@ARTICLE{Xu2023Survey,
  author={Xu, Qun-Ce and Mu, Tai-Jiang and Yang, Yong-Liang},
  journal={Computational Visual Media}, 
  title={A survey of deep learning-based 3D shape generation}, 
  year={2023},
  volume={9},
  number={3},
  pages={407-442},
  doi={10.1007/s41095-022-0321-5}}

@Article{Wu2024,
  author        = {Wu, Shuang and Lin, Youtian and Zhang, Feihu and Zeng, Yifei and Xu, Jingxi and Torr, Philip and Cao, Xun and Yao, Yao},
  title         = {Direct3D: Scalable Image-to-3D Generation via 3D Latent Diffusion Transformer},
  year          = {2024},
  month         = may,
  archiveprefix = {arXiv},
  copyright     = {arXiv.org perpetual, non-exclusive license},
  doi           = {10.48550/ARXIV.2405.14832},
  eprint        = {2405.14832},
  primaryclass  = {cs.CV},
  publisher     = {arXiv},
}

@article{li2024craftsman,
author    = {Weiyu Li and Jiarui Liu and Hongyu Yan and Rui Chen and Yixun Liang and Xuelin Chen and Ping Tan and Xiaoxiao Long},
title     = {CraftsMan3D: High-fidelity Mesh Generation with 3D Native Generation and Interactive Geometry Refiner},
journal   = {arXiv preprint arXiv:2405.14979},
year      = {2024},
}

@misc{song2021scorebased,
      title={Score-Based Generative Modeling through Stochastic Differential Equations}, 
      author={Yang Song and Jascha Sohl-Dickstein and Diederik P. Kingma and Abhishek Kumar and Stefano Ermon and Ben Poole},
      year={2021},
      eprint={2011.13456},
      archivePrefix={arXiv},
      primaryClass={cs.LG},
      url={https://arxiv.org/abs/2011.13456}, 
}

@misc{karras2022edm,
      title={Elucidating the Design Space of Diffusion-Based Generative Models}, 
      author={Tero Karras and Miika Aittala and Timo Aila and Samuli Laine},
      year={2022},
      eprint={2206.00364},
      archivePrefix={arXiv},
      primaryClass={cs.CV},
      url={https://arxiv.org/abs/2206.00364}, 
}

@Article{Zhao2024,
  author    = {Zhao, Yunkai and Wang, Lili and Zhai, Xiaoya and Han, Jiacheng and Ma, Winston Wai Shing and Ding, Junhao and Gu, Yonggang and Fu, Xiao‐Ming},
  journal   = {Advanced Science},
  title     = {Near‐Isotropic, Extreme‐Stiffness, Continuous 3D Mechanical Metamaterial Sequences Using Implicit Neural Representation},
  year      = {2024},
  issn      = {2198-3844},
  month     = nov,
  doi       = {10.1002/advs.202410428},
  publisher = {Wiley},
}

@Article{Park2024,
  author        = {Park, Jaewan and Kushwaha, Shashank and He, Junyan and Koric, Seid and Liu, Qibang and Jasiuk, Iwona and Abueidda, Diab},
  title         = {Nonlinear Inverse Design of Mechanical Multi-Material Metamaterials Enabled by Video Denoising Diffusion and Structure Identifier},
  year          = {2024},
  month         = sep,
  archiveprefix = {arXiv},
  copyright     = {Creative Commons Attribution 4.0 International},
  doi           = {10.48550/ARXIV.2409.13908},
  eprint        = {2409.13908},
  primaryclass  = {cs.AI},
  publisher     = {arXiv},
}

@Article{Ma2019,
  author    = {Ma, Wei and Cheng, Feng and Xu, Yihao and Wen, Qinlong and Liu, Yongmin},
  journal   = {Advanced Materials},
  title     = {Probabilistic Representation and Inverse Design of Metamaterials Based on a Deep Generative Model with Semi‐Supervised Learning Strategy},
  year      = {2019},
  issn      = {1521-4095},
  month     = jul,
  number    = {35},
  volume    = {31},
  doi       = {10.1002/adma.201901111},
  publisher = {Wiley},
}

@Article{Tancogne‐Dejean2018,
  author    = {Tancogne‐Dejean, Thomas and Diamantopoulou, Marianna and Gorji, Maysam B. and Bonatti, Colin and Mohr, Dirk},
  journal   = {Advanced Materials},
  title     = {3D Plate‐Lattices: An Emerging Class of Low‐Density Metamaterial Exhibiting Optimal Isotropic Stiffness},
  year      = {2018},
  issn      = {1521-4095},
  month     = sep,
  number    = {45},
  volume    = {30},
  doi       = {10.1002/adma.201803334},
  publisher = {Wiley},
}

@Article{Cheng2017,
  author    = {Cheng, Lin and Zhang, Pu and Biyikli, Emre and Bai, Jiaxi and Robbins, Joshua and To, Albert},
  journal   = {Rapid Prototyping Journal},
  title     = {Efficient design optimization of variable-density cellular structures for additive manufacturing: theory and experimental validation},
  year      = {2017},
  issn      = {1355-2546},
  month     = jun,
  number    = {4},
  pages     = {660--677},
  volume    = {23},
  doi       = {10.1108/rpj-04-2016-0069},
  publisher = {Emerald},
}

@Article{Kadic2019,
  author    = {Kadic, Muamer and Milton, Graeme W. and van Hecke, Martin and Wegener, Martin},
  journal   = {Nature Reviews Physics},
  title     = {3D metamaterials},
  year      = {2019},
  issn      = {2522-5820},
  month     = jan,
  number    = {3},
  pages     = {198--210},
  volume    = {1},
  doi       = {10.1038/s42254-018-0018-y},
  publisher = {Springer Science and Business Media LLC},
}

@article{Ipopt,
author = {W\"{a}chter, Andreas and Biegler, Lorenz T.},
title = {On the implementation of an interior-point filter line-search algorithm for large-scale nonlinear programming},
year = {2006},
issue_date = {March 2006},
publisher = {Springer-Verlag},
address = {Berlin, Heidelberg},
volume = {106},
number = {1},
issn = {0025-5610},
journal = {Math. Program.},
month = mar,
pages = {25–57},
numpages = {33}
}

@article{L-BFGS-B,
author = {Zhu, Ciyou and Byrd, Richard H. and Lu, Peihuang and Nocedal, Jorge},
title = {Algorithm 778: L-BFGS-B: Fortran subroutines for large-scale bound-constrained optimization},
year = {1997},
issue_date = {Dec. 1997},
publisher = {Association for Computing Machinery},
address = {New York, NY, USA},
volume = {23},
number = {4},
issn = {0098-3500},
url = {https://doi.org/10.1145/279232.279236},
doi = {10.1145/279232.279236},
journal = {ACM Trans. Math. Softw.},
month = dec,
pages = {550–560},
numpages = {11}
}

\clearpage
\appendix
\section*{Supplementary Materials}
\setlength{\belowcaptionskip}{4pt}
\section{Properties Calculation}
\label{sec:homogen}
We employ a GPU-accelerated implementation of the homogenization method~\cite{andreassen2014Design, dong2018149} as a solver for the properties of microstructures.
\rev{
The homogenization method computes effective macroscopic properties of heterogeneous materials by averaging local material responses across a representative volume element.
Specifically, it involves imposing prescribed macroscopic strains, solving local equilibrium equations (e.g., via finite element analysis) to determine microscale displacement and strain fields, and subsequently integrating these fields to obtain macroscopic properties.
}
Specifically, our objective is to compute the elastic tensor $E \in \mathbb{R}^{9\times9}$ of the structure.
It is derived from the fourth-order elasticity tensor $\E_{ijkl}$, which governs the linear relationship between stress $\sigma_{ij}$ and strain $\epsilon_{kl}$ as:
\begin{equation}
    \sigma_{ij} = \sum_{kl}E_{ijkl}\epsilon_{kl}.
\end{equation}
In three-dimensional space, the indices $i, j, k, l$ range from 1 to 3, resulting in $3 \times 3 \times 3 \times 3 = 81$ components in $E_{ijkl}$.
However, due to the inherent symmetries in both the stress-strain relationship (e.g., $\sigma_{ij} = \sigma_{ji}$ and $\epsilon_{ij} = \epsilon_{ji}$) and the elasticity tensor itself (e.g., $E_{ijkl} = E_{jikl} = E_{ijlk} = E_{klij}$), the number of independent components is reduced to 21.
For further simplification, we represent the elasticity tensor in Voigt notation, $\bC \in \mathbb{R}^{6\times6}$.
\rev{
For cubic symmetry, the elastic tensor exhibits significant simplification due to the high degree of symmetry in the material. In Voigt notation, the elasticity tensor $\bC$ for a cubic material reduces to a form with only three independent constants: $C_{11}$, $C_{12}$, and $C_{44}$. The resulting $6 \times 6$ matrix representation is given by:
\begin{equation}
\bC =
\begin{bmatrix}
C_{11} & C_{12} & C_{12} & 0 & 0 & 0 \\
C_{12} & C_{11} & C_{12} & 0 & 0 & 0 \\
C_{12} & C_{12} & C_{11} & 0 & 0 & 0 \\
0 & 0 & 0 & C_{44} & 0 & 0 \\
0 & 0 & 0 & 0 & C_{44} & 0 \\
0 & 0 & 0 & 0 & 0 & C_{44}
\end{bmatrix}.
\end{equation}
}

The components of $\bE_{ijkl}$ are computed as follows:
\begin{equation}
\label{eq:E}
    E_{ijkl} = \frac{1}{|\bV|} \int_\bV \left( \overline{\epsilon}_{ij} - \epsilon_{ij}(\boldsymbol{\chi}) \right)  E^b_{ijkl} \left(\overline{\epsilon}_{kl} - \epsilon_{kl}(\boldsymbol{\chi}) \right) d\bV,
\end{equation}
where $\bE^b$ denotes the locally varying elasticity tensor, $\bV$ is the unit cell with volume $|\bV|$, $\overline{\epsilon}$ represents the prescribed macroscopic strain fields, and $\epsilon(\boldsymbol{\chi})$ denotes the locally varying strain fields.
The displacement fields $\boldsymbol{\chi}$, introduced during the homogenization process, encode critical physical priors but are also the most computationally expensive component to calculate.

The calculation of $\boldsymbol{\chi}$ can be formulated as solving the following equation :
\begin{equation}
\label{eq:chi_eq}
    \int_\bV E_{ijkl} \epsilon_{ij}(v)\epsilon_{pq}(\boldsymbol{\chi}) d\bV = \int_\bV E_{ijkl} \epsilon_{ij}(v) \overline{\epsilon}_{kl} d\bV \quad \forall v \in \bV,
\end{equation}
where $v$ is a virtual displacement field.

Numerically, this equation is solved using the finite element method (FEM), where the domain $\bV$ is discretized into a grid with $n$ resolution.
For each element $e$, the local stiffness matrix $\bK_e \in \mathbb{R}^{24\times24}$ and load vector $\bff_e \in \mathbb{R}^{24\times6}$ are constructed.
These local contributions are then assembled into the global stiffness matrix $\bK$ and the global load vector $\bF$ over all valid elements, leading to a global system of linear equations:
\begin{equation}
    \bK\boldsymbol{\chi} = \bF,
\end{equation}
which serves as the foundation for incorporating physical priors into our formulation, as referenced in Sec.~\ref{sec:phy_enc}.

In the FEM context, Eq.~\ref{eq:E} is performed by:
\begin{equation}
        E_{ijkl} = \frac{1}{|\bV|} \sum_{v \in \bV} \left( \overline{\epsilon} - \epsilon(\boldsymbol{\chi}) \right)^T  E^b_{ijkl} \left(\overline{\epsilon} - \epsilon(\boldsymbol{\chi}) \right).
\end{equation}

The mechanical properties, including Young’s modulus ($E$), Poisson’s ratio ($\nu$), and shear modulus ($G$), can be derived from the elastic tensor $\bC$.
First, the compliance matrix $\bS$ is obtained by inverting the elastic tensor:
\begin{equation}
\bS = \bC^{-1}.
\end{equation}
For microstructures with cubic symmetry, the mechanical properties can be expressed as:
\begin{equation}
    E=\frac{1}{S_{11}}, \nu = -\frac{S_{12}}{S_{11}}, G=\frac{1}{S_{44}}.
\end{equation}

\section{Implementation details}
\label{sec:imple_detail}
\rev{
We employ a 7-layer residual convolutional neural network (CNN) with $3\times3$ kernels to encode the Holoplane.
During decoding, after projecting any point $\bx$ onto the Holoplane to obtain $\bu$, we feed $\holo(\bu)$ into a 5-layer residual MLP to predict the corresponding fields.
}

We train the Holoplane autoencoder using the SDF of the mesh voxelized at a resolution of $128^3$ as input.
\rev{And we sample 100,000 points within the SDF randomly and another 100,000 points near the mesh surface as references.}
We compute the displacement field $\chi_\Omega$ for each microstructure using the homogenization solver.
All training procedures were conducted on a Linux server equipped with four NVIDIA A40 GPUs.
The autoencoder was trained for approximately 72 hours, and the diffusion model required around 120 hours of training.

\rev{
In our experiments, we focus on cubic symmetry and thus employ only one of the three orthogonal symmetry planes for encoding. We simultaneously sample $\holo(\x, y)$ and $\holo(y, x)$, and their average is provided as input to the decoder. This strategy ensures that the generated structures strictly conform to the 48 symmetry operations defined by the octahedral group $O_h$.
}

\rev{
For other symmetry groups, such as tetrahedral, all three orthogonal planes can be utilized, while omitting the sampling of $\holo(y, x)$. This configuration effectively removes rotational symmetries while preserving only mirror-plane symmetries, thereby reducing the group to 8 symmetry operations. The choice of sampling strategy and plane configuration directly determines the symmetry constraints imposed on the generated structures.
}

The minimum and maximum values in our dataset used for error calculation in Sec.~\ref{sec:accuracy} are provided in Tab.~\ref{tab:min_max}.
\begin{table}[ht]
    \centering
    \caption{Minimum and maximum values of the elastic constants ($C_{11}$, $C_{12}$, and $C_{44}$)}
    \label{tab:min_max}
    \begin{tabular}{lccc}
        \toprule
                        & C\textsubscript{11} & C\textsubscript{12} & C\textsubscript{44} \\
        \midrule
        Min             & 0.0004 & -0.0078 & 0.0000 \\
        Max             & 0.6209 & 0.2368 & 0.1591 \\
        \bottomrule
    \end{tabular}
\end{table}

\section{Heterogeneous design details}
\label{sec:heterogeneous}
With MIND, we can input material properties into the network to generate corresponding microstructures.
However, for a given input design, we must determine the distribution of material properties that best approximates the specified target behavior.
To achieve this, we implement a linear finite element method (FEM) based on hexahedron discretization to simulate the behavior of a given design. Using the regular hexahedron mesh, we compute the static equilibrium state by solving the following minimization problem:
\begin{equation}
    \argmin_\bu \frac{1}{2}\bu^\intercal \bK \bu - \bff_{ext}^\intercal \bu \ ,
\end{equation}
where $\bu$ represents the nodal displacements, $\bff_{ext}$ denotes the applied external forces, and $\bK$ is the global stiffness matrix evaluated from the material properties $\mathbf{C}_i$ of each element.
To handle fixed vertices, we filter the Hessian and gradient for the corresponding (DoFs) prior to solving the linear system. This minimization problem is solved using a standard Newton's solver. With this setup, our neural networks can seamlessly incorporate material properties derived from this computational model.

\paragraph{Optimizing material properties} The computational model allows us to predict the behaviors of a given design with specified material properties. In order to find material properties that best approximate the target behavior, we optimize material properties by solving the following inverse problem,
\begin{equation}
    \argmin_\bp T_u = \sum_k\|\bu_k(\bp_s) - \hat{\bu}_k\|^2 \quad\text{s.t.}\quad  \bff(\bp_s) = \mathbf{0},
    \label{eq:designobjective_nolimiting}
\end{equation}
where $\bu_k$ and $\hat{\bu}_k$ are the deformed and target displacements for specified vertices, $\bp_s = S^L(\bp)$ is the smoothed material properties via Laplacian smoothing operations $S$.
The above design objective allows us to compute the optimal material properties that minimize the difference to the target behavior.
Even though we can solve the above optimization problem with bounds constraints for $\bp$, the optimized material properties might be outside of the feasible region of the network capabilities. Similar to \cite{tang2023beyond}, we represent the feasible space of the three-dimensional material property space $\mathbf{C}$ as a triangle mesh $\ssm$.  We can leverage collision detection algorithms to constrain the material properties. To this end, we interpret the material properties $\mathbf{C}_i$ of each FEM element as a point in three-dimensional material space. Instead of enforcing points strictly within the mesh of material space, we use soft constraints that penalize the points in material space outside the mesh. We first compute the distance $d_i$ of $\bs_i$ to the closest primitive---point, edge, or triangle---on $\ssm$ using standard geometry tests and bounding volume hierarchies for acceleration. We then set up smooth, unilateral penalty functions as
\begin{equation}
    T^m_i = \begin{cases}
    k_m d_i^2 & d_i > 0 \\
    0 & d_i\leq 0
    \end{cases} \ ,
    \label{eq:materialSpaceLimiting}
\end{equation}
where $k_m$ is the penalty stiffness. Since we do not have to strictly enforce the material properties within the boundary of the feasible material space, we find a soft stiffness with a value of $k_m=1.0$ is suitable for all the cases. By combining (\ref{eq:designobjective_nolimiting}) and (\ref{eq:materialSpaceLimiting}), we obtain the design objective for optimizing feasible material properties
\begin{equation}
    \argmin_\bp T(\bp) = T_u(\bp) + \sum_i T_i^m(\bp)
    \quad\text{s.t.}\quad \bff(\bp_s) = \mathbf{0} \ .
    \label{eq:materialOpt}
\end{equation}
We solve this design objective by casting it as an unconstrained optimization problem based on sensitivity analysis of the static equilibrium state. Based on the computed design objective gradient from the sensitivity matrix, we solve this problem via L-BFGS-B~\cite{L-BFGS-B} with bound constraints to enforce the physical meaning of material properties. Note that we optimize the material properties $\bp$ in the design objective, however, we use the smoothed material properties $\bp_s$ to generate microstructures from neural networks.

For cases with limited resources or requiring weight control, we further develop a design objective to maximize material utilization for a specified target,
\begin{equation}
    \begin{gathered}
    \argmin_\bp T(\bp) = T_u(\bp) + \sum_i T_i^m(\bp)
    \\ \quad\text{s.t.}\quad \bff(\bp_s) = \mathbf{0}  \quad\text{and}\quad \sum_i E_i = b \ ,
    \label{eq:materialOpt_budget}
    \end{gathered}
\end{equation}
where we use Young's modulus as the cost for each element. We solve this problem by using Ipopt~\cite{Ipopt} with bounds constraints for material properties.

\section{Visualization of generated Results}
\label{sec:more_vis}
\begin{figure*}[tb]
    \centering
    \includegraphics[width=\linewidth]{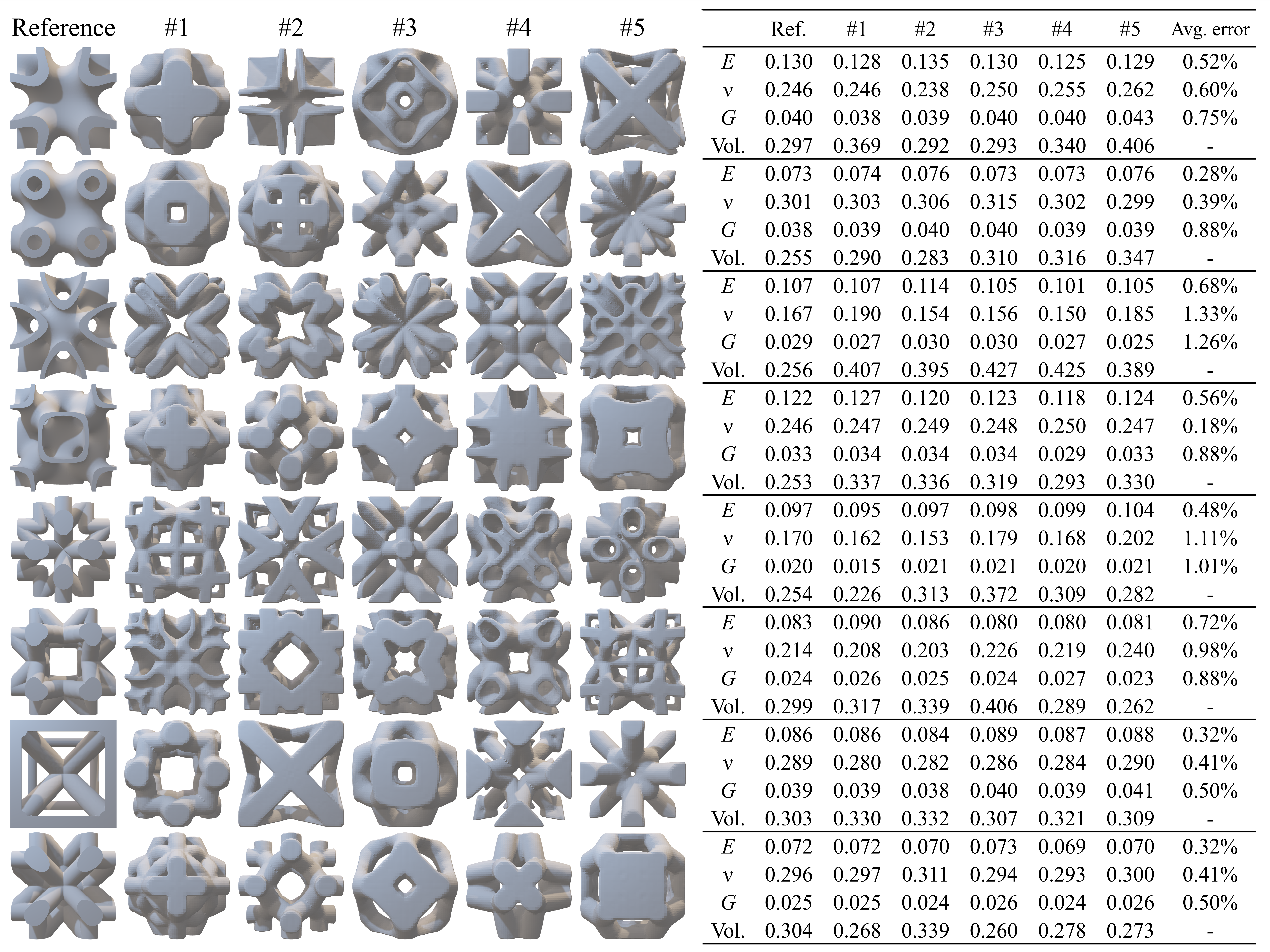}
    \caption{More generated results from a reference model.
Using the reference model's mechanical properties as input, five candidate models generated by our framework are shown. The properties values ($E,\nu,G$) for both the reference model and generated models are listed. }
    \label{fig:supp_reference}
\end{figure*}

We visualize additional generated results and provide their corresponding mechanical properties as reference (Fig.~\ref{fig:supp_reference}).

\end{document}